\documentclass{article}

\usepackage{microtype}
\usepackage{graphicx}
\usepackage{subcaption}
\usepackage{booktabs} 

\usepackage{hyperref}

\usepackage[preprint]{icml2026}

\usepackage{amsmath}
\usepackage{amssymb}
\usepackage{mathtools}
\usepackage{amsthm}

\usepackage[capitalize,noabbrev]{cleveref}

\theoremstyle{plain}

\theoremstyle{definition}

\theoremstyle{remark}

\usepackage{afterpage}

\usepackage[textsize=tiny]{todonotes}

\icmltitlerunning{Cheap2Rich for Data Assimilation}

\begin{document}

\twocolumn[
  \icmltitle{Cheap2Rich: A Multi-Fidelity Framework for Data Assimilation and System Identification of Multiscale Physics - Rotating Detonation Engines}






  \icmlsetsymbol{equal}{*}

  \begin{icmlauthorlist}
    \icmlauthor{Yuxuan Bao}{equal,amath}
    \icmlauthor{Jan Zajac}{equal,ece,eth}
    \icmlauthor{Megan Powers}{umich}
    \icmlauthor{Venkat Raman}{umich}
    \icmlauthor{J. Nathan Kutz}{amath,ece}

  \end{icmlauthorlist}

  \icmlaffiliation{ece}{Department of Electrical and Computer Engineering, University of Washington, Seattle, USA}
  \icmlaffiliation{amath}{Department of Applied Mathematics, University of Washington, Seattle, USA} 
  \icmlaffiliation{umich}{University of Michigan, Advanced Propulsion Concepts Lab, Ann Arbor, USA}
  \icmlaffiliation{eth}{Department of Mathematics, Swiss Federal Institute of Technology Zurich, Zurich, Switzerland}

  \icmlcorrespondingauthor{J. Nathan Kutz}{kutz@uw.edu}

  \icmlkeywords{Machine Learning, ICML, Rotating detonation engines, data assimilation, SHRED, latent dynamics, surrogate modeling, injector physics}

  \vskip 0.3in
]



\printAffiliationsAndNotice{\icmlEqualContribution}

\begin{abstract}
    Bridging the sim2real gap between computationally inexpensive models and complex physical systems remains a central challenge in machine learning applications to engineering problems, particularly in multi-scale settings where reduced-order models typically capture only dominant dynamics. 
    In this work, we present Cheap2Rich, a multi-scale data assimilation framework that reconstructs high-fidelity state spaces from sparse sensor histories by combining a fast low-fidelity prior with learned, interpretable discrepancy corrections. 
    We demonstrate the performance on rotating detonation engines (RDEs), a challenging class of systems that couple detonation-front propagation with injector-driven unsteadiness, mixing, and stiff chemistry across disparate scales. 
    Our approach successfully reconstructs high-fidelity RDE states from sparse measurements while isolating physically meaningful discrepancy dynamics associated with injector-driven effects.
    The results highlight a general multi-fidelity framework for data assimilation and system identification in complex multi-scale systems, enabling rapid design exploration and real-time monitoring and control while providing interpretable discrepancy dynamics.
    Code for this project is is available at: \url{github.com/kro0l1k/Cheap2Rich}.
\end{abstract}


\begin{figure}[t]
  \centering

  \begin{subfigure}{\linewidth}
    \centering
    \includegraphics[width=0.98\linewidth]{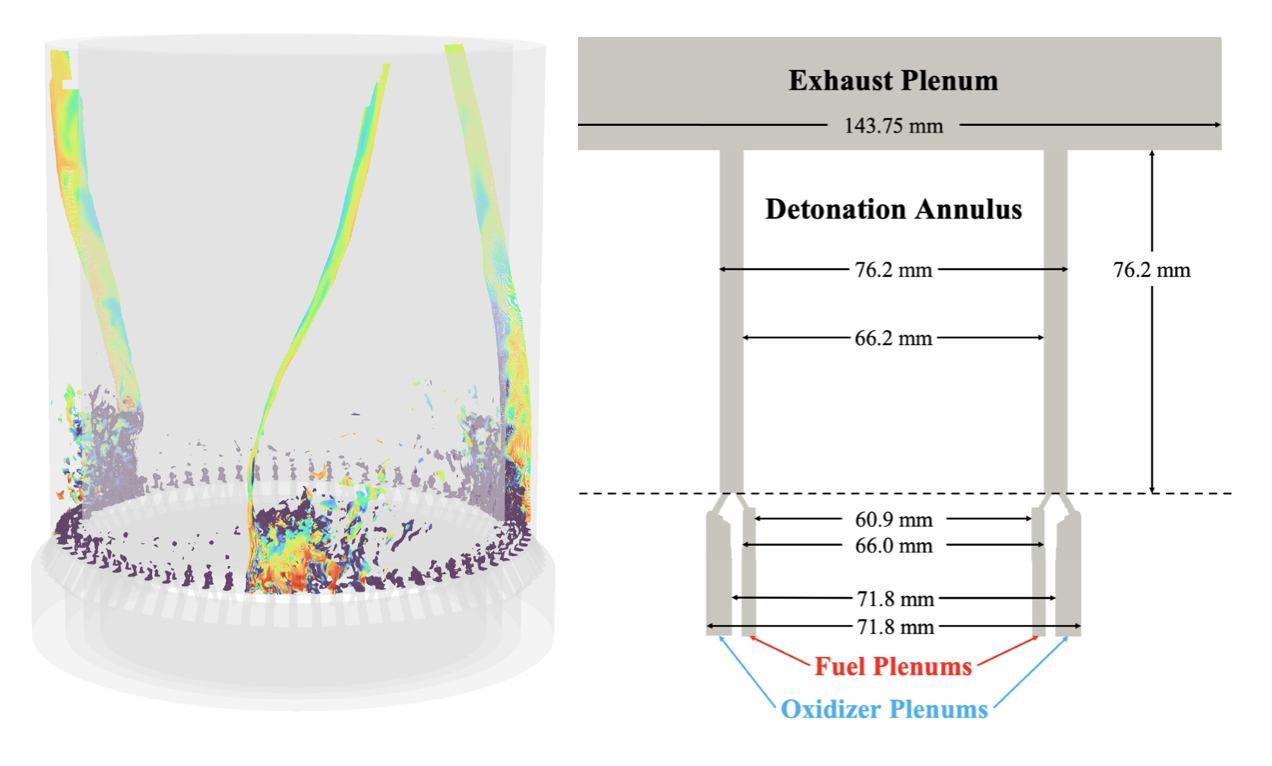}
    \label{fig:3d_description}
  \end{subfigure}

  \vspace{-0.6em} 

  \begin{subfigure}{\linewidth}
    \centering
    \includegraphics[width=0.98\linewidth]{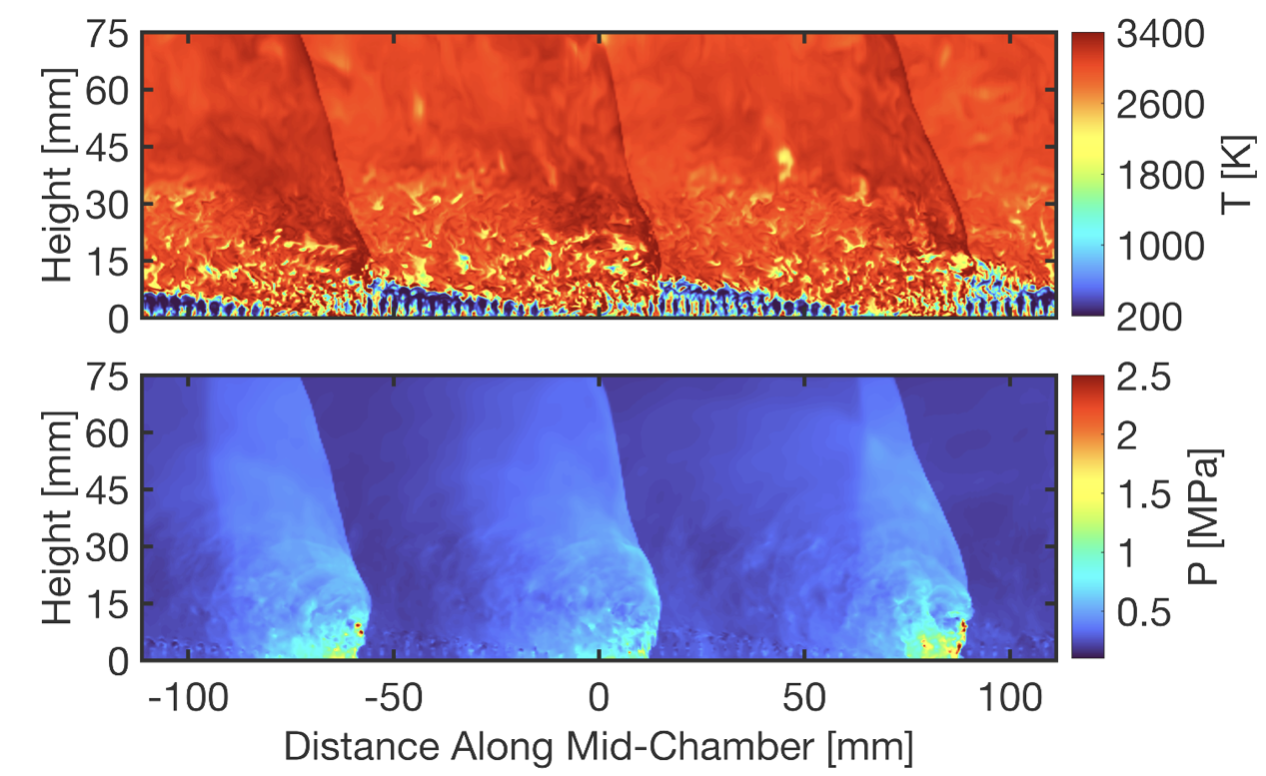}
    \label{fig:2d_profile}
  \end{subfigure}

  \label{fig:sim_geometry_and_profiles}
  \caption {A rotating detonation rocket engine (RDRE). Full-scale geometry (top left) and cross-sectional schematic with primary dimensions (top right)  Temperature and Pressure mid-channel contour projections (bottom).}
\label{fig:3d_description_full}
\end{figure}

\section{Introduction}

\begin{figure*}[t]
\centering
\includegraphics[width=0.85\textwidth]{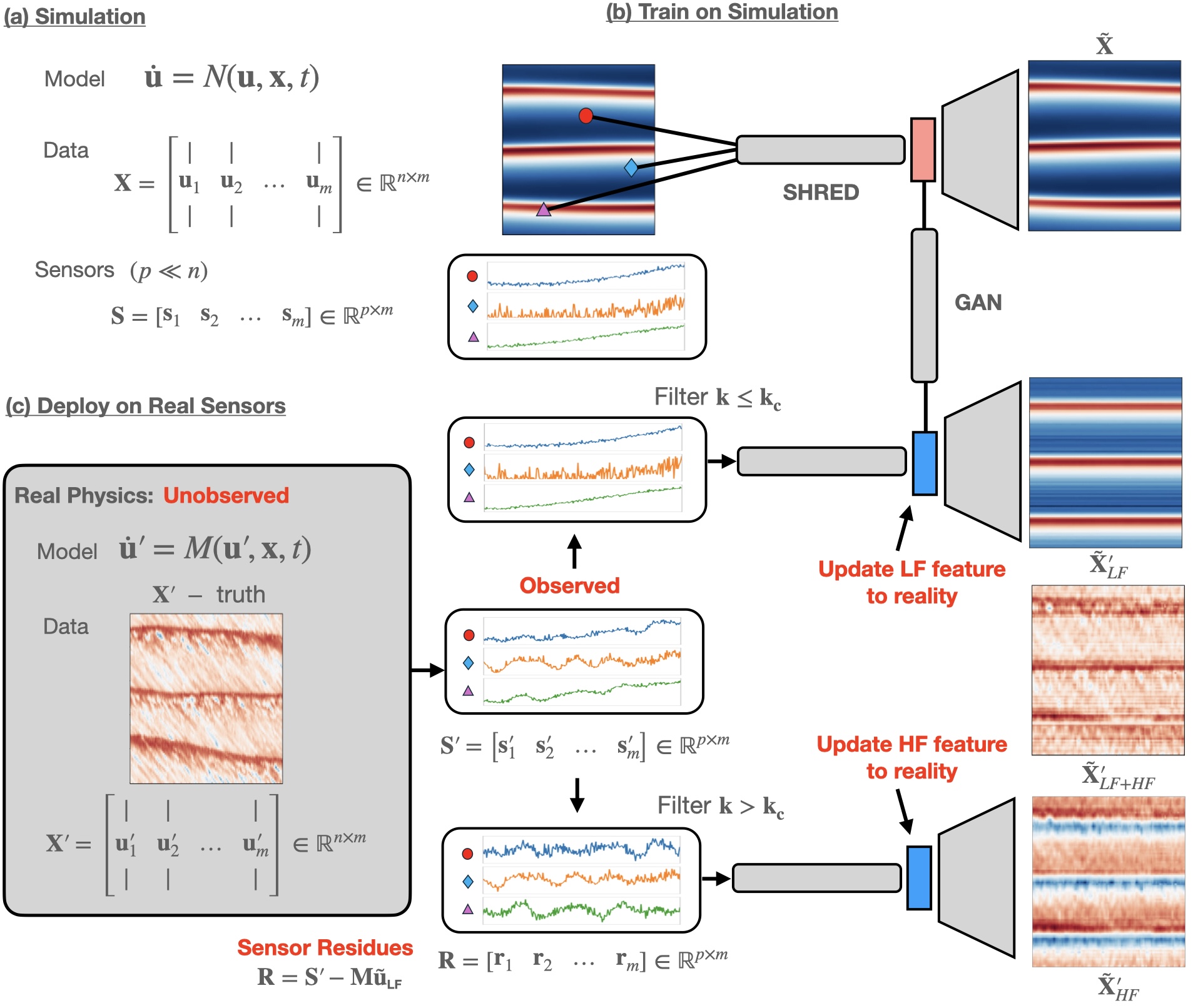}
\caption{The Cheap2Rich architecture. (a) Simulation model provides full-state data $\mathbf{X}$ and sparse sensor measurements $\mathbf{S}$. (b) A standard SHRED network is trained on simulation data to reconstruct full states $\tilde{\mathbf{X}}$ from sensor histories. (c) Deployment on real sensors: the real physics full state $\mathbf{X}'$ is unobserved; only sparse sensor measurements $\mathbf{S}'$ are available. The LF DA-SHRED pathway updates the latent features to reality, producing $\tilde{\mathbf{X}}'_{\text{LF}}$. The HF pathway processes sensor residuals $\mathbf{R}$ to capture fine-scale corrections $\tilde{\mathbf{X}}'_{\text{HF}}$. The final reconstruction $\tilde{\mathbf{X}}'_{\text{LF+HF}}$ combines both pathways to close the simulation-to-reality gap.}
\label{fig:architecture_overview}
\end{figure*}

Despite significant advancements in physics-informed AI~\cite{kutz2025accelerating, karniadakis2021physics,brunton2022data,wyder2025common,fan2025physics} and neural operators~\cite{lu2021learning, li2020fourier, roy2025physics}, modeling of multi-scale physics remains a grand challenge due to the inability of methods to model the  orders of magnitude different time and space scales present in complex systems. Traditional machine learning methods typically capture the dominant large scale time and space features due to the spectral bias in training~\cite{rahaman2019spectral}.  Thus important characteristics and features are effectively band-pass filtered in model training.  We introduce a neural network architecture that by construction targets different scales for training. Specifically, Data assimilation (DA) \cite{DA_ocean, da_review} provides a principled framework for closing the performance gap in multiscale modeling by combining physics-based prediction and experimental observation. Recent advances in latent dynamical system learning~\cite{erichson2020shallow, jiang2025hierarchical}, including the Shallow Recurrent Decoder (SHRED) framework \cite{sensing_with_shred,tomasetto2025reduced,bao2025data}, further enable the discovery of compact latent representations and missing functional structure directly from data. However, the application of DA-enhanced latent learning to strongly nonlinear, multistable systems with sparse, noisy measurements remains largely unexplored \cite{niu2024multi,liu2024kan}.

In this work, we introduce a multi-scale data assimilation pipeline that bridges a fast reduced-order RDE model and high-fidelity dynamics using only sparse sensor histories. The method augments a SHRED framework trained on multiscale data with a deployment-time assimilation mechanism that separates the reconstruction into (i) a low-frequency (LF) component, aligned to the sensor-induced latent distribution via a lightweight latent-GAN and explicitly low-pass filtered, and (ii) a high-frequency (HF) correction learned directly from sensor-space residuals and regularized to be spectrally sparse. This decomposition enforces a clean division between dominant features that the reduced model already captures and fine-scale discrepancies induced by unmodeled or unresolved physics, yielding a unified surrogate that is both accurate and diagnostically informative.

Rotating detonation engines (RDEs), which are an ideal test case for our method due to the inherent multiscale physics, offer a promising pathway toward high-efficiency propulsion and power generation by sustaining continuous detonation waves within an annular combustor \cite{Mixing_and_detonation_RDE, Space_Flight_of_RDE, Experimental_Visualization_of_RDEs, Detonative_propulsion_review}. However, the underlying physics of RDEs is governed by tightly coupled compressible flow, shock-detonation interactions, stiff chemical kinetics, and injector-driven mass, momentum, and energy exchange across disparate spatial and temporal scales \cite{VenkatNonlinearities_RDE}. High-fidelity numerical simulations capable of resolving these coupled processes are computationally prohibitive, often requiring weeks of supercomputing time for a single operating condition \cite{Megan_RDE}. This computational burden fundamentally limits systematic design exploration, uncertainty quantification, and control-oriented modeling.
To address these challenges, reduced-order physics-based models have been developed to capture the dominant detonation-front dynamics at dramatically reduced cost \cite{Koch_multiscale_Physcics, KochRDE, KochDD_RDE}. Among these, the one-dimensional rotating detonation model of Koch and collaborators has emerged as a widely used surrogate for azimuthal detonation propagation \cite{KochRDE}. While such models accurately represent leading-order detonation physics, they necessarily omit or heavily simplify injector dynamics, mixing processes, time-delay effects, and non-equilibrium losses. As a result, substantial and systematically structured discrepancies persist between low-order model predictions and real experimental RDE data.  We show that we can leverage Koch's cheap simulation model (minutes) to approximate simulations of the rich model (weeks), thus allowing for a Cheap2Rich algorithm whereby cheap proxies can be used to model the exceptionally rich multiscale physics observed in reality.  

On a three-wave co-rotating configuration (See Fig.~\ref{fig:architecture_overview}), the learned LF model remains close to the simple Koch's model prior and cannot reproduce injector-driven variability.   Adding a learned HF pathway reduces the error (an $80.9\%$ reduction) while exhibiting sparse Fourier content concentrated at harmonics of the three-wave structure, indicating phase-locked corrections tied to detonation-front passage and consistent with injector-modulated forcing absent in the baseline model. This yields a practical Cheap2Rich bridge: the reduced model runs in seconds, and the trained network performs full-state reconstruction from sparse measurements at negligible marginal cost compared to high-fidelity simulation, while the structured HF term provides direct access to the subsequent missing-physics identification (e.g., via SINDy~\cite{brunton2016discovering}). 

\section{Preliminaries}

\subsection{RDEs - high fidelity simulation}

There has been a lot of prior work on accurate modeling of RDEs that stem from high-fidelity numerical simulations \cite{VenkatNonlinearities_RDE}. In this work, a high-fidelity simulation of the AFRL methane-oxygen rotating detonation rocket engine (RDRE) was studied as shown in Fig.\ref{fig:3d_description_full}. This geometry was chosen due to the high availability of experimental \cite{bennewitz2019modal} and numerical simulations \cite{prakash2021numerical} available. The reacting compressible Navier-Stokes and species transport equations \eqref{eq:gov_eqs} are solved using an in-house compressible flow solver \cite{numericsHMM} built on the AMReX library \cite{AMReX}, which provides the adaptive mesh refinement (AMR) framework. 

\afterpage{
\begin{equation}
\label{eq:gov_eqs}
\begin{aligned}
&\frac{\partial \rho}{\partial t}+\frac{\partial(\rho u_i)}{\partial x_i}=0,\\[1pt]
&\frac{\partial (\rho u_i)}{\partial t}+\frac{\partial(\rho u_i u_j)}{\partial x_j}
=-\frac{\partial p}{\partial x_i}+\frac{\partial \tau_{ij}}{\partial x_j},\\[1pt]
&\frac{\partial (\rho e)}{\partial t}+\frac{\partial(\rho h\,u_j)}{\partial x_j}
=\frac{\partial}{\partial x_j}\!\left(\alpha \frac{\partial T}{\partial x_j}\right)
+\frac{\partial(u_i\tau_{ij})}{\partial x_j}
 \\
& \qquad \qquad \qquad +  \sum_{k=1}^{N_s} h_k\,\frac{\partial}{\partial x_j}\!\left(\rho D \frac{\partial Y_k}{\partial x_j}\right),\\[1pt]
&\frac{\partial (\rho Y_k)}{\partial t}+\frac{\partial(\rho Y_k u_j)}{\partial x_j}
=\frac{\partial}{\partial x_j}\!\left(\rho D \frac{\partial Y_k}{\partial x_j}\right)+\Omega_k
\end{aligned}
\end{equation}
}

for $k=1,\dots,N_s$ where $\rho$ is the density of the fluid, t is time, $x_i$ and $u_i$ are the spatial coordinate and the velocity component in the $i^{t h}$ direction, respectively. The viscous stress tensor is given as $\tau_{i j}$ and p is the fluid pressure. The total chemical energy is defined as $\mathrm{e}, \mathrm{h}$ is the total enthalpy of the mixture, $\alpha$ is the thermal conductivity of the mixture, D is the diffusivity, and T is the temperature. The mass fraction and chemical source term for the $k^{t h}$ species are given by $Y_k$ and $\Omega_k$, respectively. These equations are solved in a Cartesian coordinate system using a second-order finite-volume discretization method,
and time integration is performed using a strong stability-preserving second-order Runge–Kutta scheme. Detailed finite-rate chemistry is simulated using Cantera with the FFCMy-12 mechanism \cite{FoundationalFuelChem, FFCMY_personal} consisting of 12 species and 38 reactions. For additional details on the nnumerics and implementation, the reader is referred to \cite{numericsHMM,GPU-solver-for-RDEs,bielawski2023highly}.

The high-fidelity simulation employs a nonuniform grid. A fine resolution of 93.6 $\mu$m is used in the injectors, plenums, lower half of the combustion chamber, and in the regions containing the detonation wave. The grid is coarsened to 374 $\mu$m in the upper half of the combustion chamber and further to 748 $\mu$m in the exhaust plenum. The detonation wave is dynamically tagged using the same pressure gradient threshold as \cite{Megan_RDE}. The resulting cell count for the simulation was 178 million cells and was run using 3500 CPUs. The simulation cost more than 2 million CPU hours and ran for over 2 months. This simulation required 3.2 billion degrees of freedom.

To ignite the RDRE, four high temperature/pressure kernels were equally spaced in the domain and allowed to evolve. After 0.7 ms the simulation reached steady-state and produced three co-rotating waves. The simulation was allowed to continue for additional 0.5 ms of data collection for averaging. Figure~\ref{fig:3d_description_full} shows the unwrapped view that highlights the temperature and pressure contours. Also, the dynamic tagging of the detonation waves are shown in the 3D geometry. 250 snapshots are considered for the following analysis which corresponds to one full rotation of the waves in the detonation annulus. We then project the AMR results first onto a fixed 3d cylindrical mesh with $30000$ points and then on a 1d ring with $100$ points at the distance from injectors of $20 mm$ (See  Appendix \ref{app:preprocessing}).

\begin{figure}[t]
  \centering
  \includegraphics[width=0.49\linewidth]{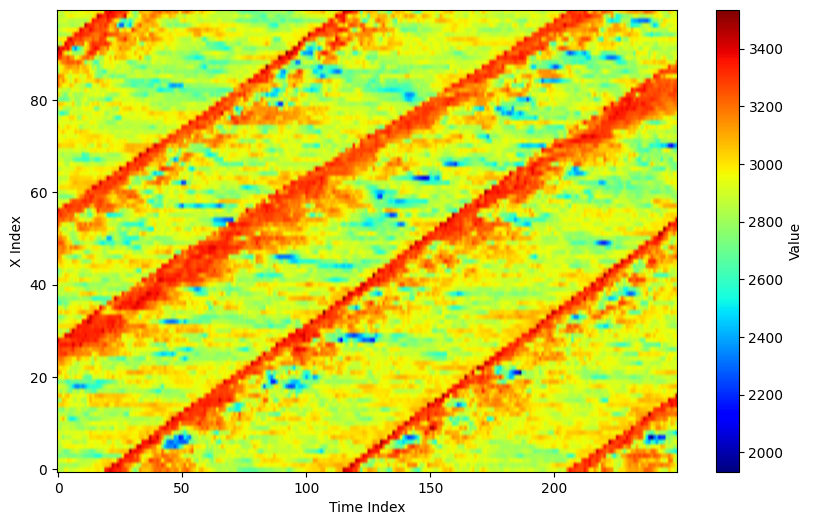}
  \hfill
  \includegraphics[width=0.49\linewidth]{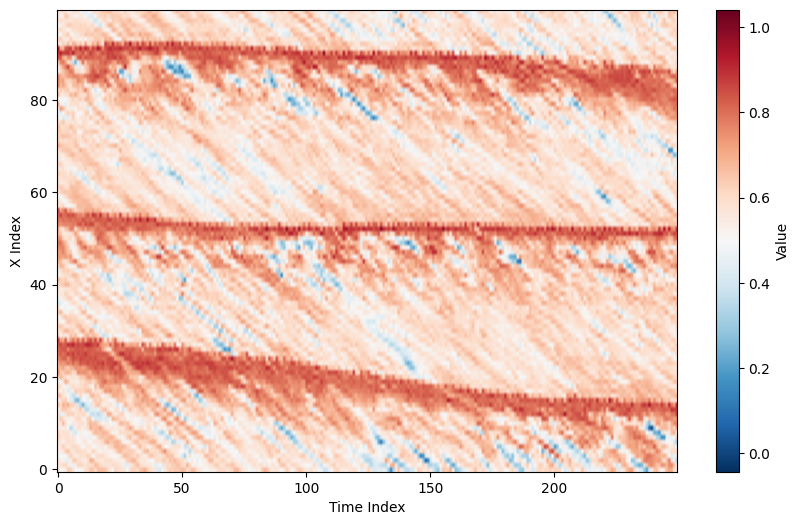}
  \caption{Time evolution of 1d projection of temperature contour before preprocessing (left), and in the COM frame of reference after min-max rescaling (right)}
  \label{fig:two-panels}
\end{figure}

\subsection{Koch's One-Dimensional RDE Model}
\label{subsec:description_of_Koch}

\begin{figure}[t]
  \centering
  \begin{subfigure}[b]{0.49\linewidth}
    \centering
    \includegraphics[width=\linewidth]{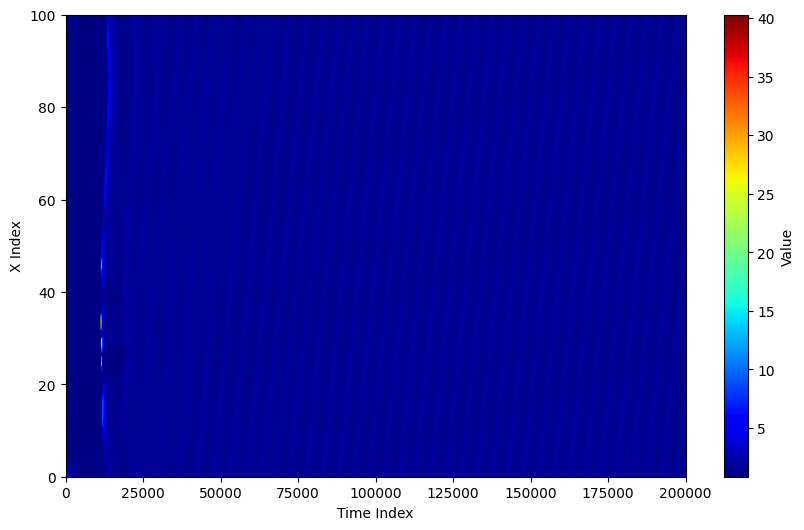}
  \end{subfigure}
  \hfill
  \begin{subfigure}[b]{0.49\linewidth}
    \centering
    \includegraphics[width=\linewidth]{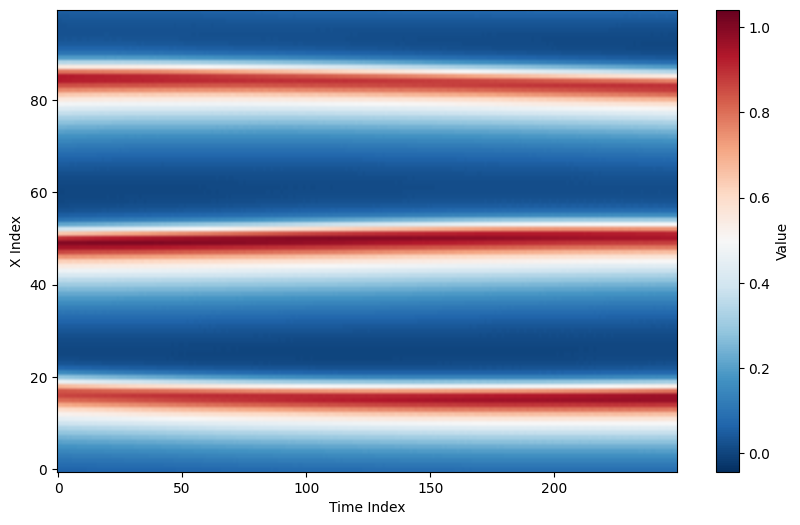}
  \end{subfigure}
\caption{Temperature field obtained from Koch's model described in Subsection \ref{subsec:description_of_Koch} before (left) and after (right) preprocessing. }
\label{fig:Koch1d}
\end{figure}

The model described in \cite{KochRDE} is a 1D model of the reactive Euler equations with source terms representing injection, mixing, and chemical kinetics inside an RDE combustion chamber. We describe the model in detail in the appendix \ref{app:KochModel}. With the set of parameters described in \ref{tab:assim_params} it is able to model the three co-rotating waves we observe in the high-fidelity simulation. We select time indices $[100000, 110000]$ which correspond to roughly one rotation of the three waves around the ring, and preprocess them by first subsampling them by taking every 400-th snapshot, leaving exactly 250 timesteps which exactly corresponds to the setting of the high-fidelity simulation. We then cancel out the rotation and scale the temperature field to $[0,1]$ linearly, obtaining the spatio-temporal field in Fig.~\ref{fig:Koch1d}.

\section{Cheap2Rich Pipeline}

We propose a multi-scale data assimilation framework that decomposes the full-state reconstruction into an additive low-frequency (LF) backbone and a spectrally-constrained high-frequency (HF) residual. This Cheap2Rich architecture leverages the SHRED and DA-SHRED framework \cite{sensing_with_shred, bao2025data} to bridge the simulation-to-reality gap by separating the dominant dynamics captured by simplified models from the fine-scale corrections that account for missing physics. The key insight is that enforcing spectral sparsity on the HF correction encourages physically interpretable, parsimonious representations of the discrepancy~\cite{brunton2016discovering}.

\subsection{Problem Setup}

\begin{figure}[t]
\centering
\resizebox{0.95\columnwidth}{!}{%
\begin{tikzpicture}[node distance=0.9cm, auto,
    block/.style={rectangle, draw, text width=1.8cm, text centered, minimum height=0.7cm, font=\scriptsize},
    arrow/.style={->, >=stealth, thick}]
    
    \node[block] (input) {Sensor History $\mathbf{S}'_{t-L:t}$};
    
    \node[block, below left=1.0cm and 0.3cm of input] (lf_lstm) {LF-LSTM Encoder};
    \node[block, below=0.6cm of lf_lstm] (gan) {Latent GAN};
    \node[block, below=0.6cm of gan] (lf_decoder) {LF Decoder};
    
    \node[block, below right=1.0cm and 0.3cm of input] (residual) {Residual Comp.};
    \node[block, below=0.6cm of residual] (hf_lstm) {HF-LSTM Encoder};
    \node[block, below=0.6cm of hf_lstm] (hf_decoder) {HF Decoder + Deform};
    
    \node[block, below=2.8cm of input, yshift=-1.5cm] (output) {$\tilde{\mathbf{u}}' = \tilde{\mathbf{u}}_{\text{LF}} + \tilde{\mathbf{u}}_{\text{HF}}$};
    
    \draw[arrow] (input) -- (lf_lstm);
    \draw[arrow] (input) -- (residual);
    \draw[arrow] (lf_lstm) -- (gan);
    \draw[arrow] (gan) -- (lf_decoder);
    \draw[arrow] (lf_decoder) -- (output);
    \draw[arrow] (lf_decoder.east) -- ++(0.3,0) |- (residual.west);
    \draw[arrow] (residual) -- (hf_lstm);
    \draw[arrow] (hf_lstm) -- (hf_decoder);
    \draw[arrow] (hf_decoder) -- (output);
    
\end{tikzpicture}%
}
\caption{Schematic of the Cheap2Rich architecture. The LF pathway learns the dominant dynamics from simulation and aligns to reality via a latent GAN. The HF pathway learns spectrally-sparse corrections from sensor residuals.}
\label{fig:architecture}
\end{figure}
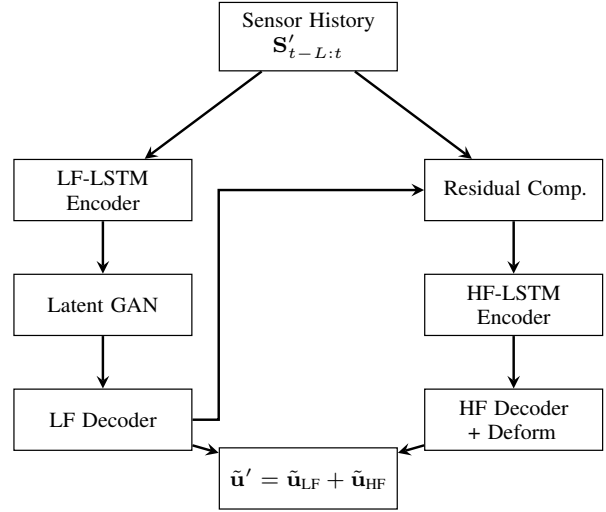

We summarize the notation used throughout this section in Table~\ref{tab:notation} (Appendix~\ref{app:notation}), following the conventions established in \cite{bao2025data}.
The full-state reconstruction is decomposed as
{
\setlength{\abovedisplayskip}{4pt}
\setlength{\belowdisplayskip}{4pt}
\begin{equation}
\tilde{\mathbf{u}}'(t) = \tilde{\mathbf{u}}_{\text{LF}}(t) + \tilde{\mathbf{u}}_{\text{HF}}(t),
\label{eq:decomposition}
\end{equation}
}where $\tilde{\mathbf{u}}_{\text{LF}}$ captures the dominant low-frequency structure learned from simulation and adapted to reality via latent-space alignment, and $\tilde{\mathbf{u}}_{\text{HF}}$ represents a spectrally-sparse high-frequency correction.
The Cheap2Rich architecture consists of two parallel pathways that process the sensor time-history $\{\mathbf{s}'_{t-\ell}\}_{\ell=0}^{L-1}$ with $L$ temporal lags, as illustrated in Figure~\ref{fig:architecture}.

\subsection{Low-Frequency Pathway}

The LF pathway follows the standard DA-SHRED methodology \cite{bao2025data}, consisting of a temporal encoder trained on simulation data and a latent-space alignment.

\subsubsection{Temporal Encoder}

Given sensor history $\mathbf{S}'_{t-L:t} = [\mathbf{s}'_{t-L+1}, \ldots, \mathbf{s}'_t] \in \mathbb{R}^{L \times p}$, the LF encoder maps this sequence to a latent representation with a two-layer LSTM \cite{hochreiter1997long}:
\begin{equation}
\mathbf{z}_{\text{LF}}(t) = \mathcal{E}_{\text{LF}}(\mathbf{S}'_{t-L:t}; \boldsymbol{\theta}_{\text{enc}}) = \text{LayerNorm}\left(\mathbf{h}^{(2)}_L\right),
\label{eq:lf_encoder}
\end{equation}
where $\mathbf{h}^{(2)}_L \in \mathbb{R}^{d_z}$ is the final hidden state of the second LSTM layer with hidden dimension $d_z$, and LayerNorm denotes normalization for training stability \cite{ba2016layer}.

\subsubsection{Latent-Space Alignment via GAN}

To bridge the distribution shift between simulation-trained latent codes and those induced by real sensor measurements, we employ a residual generator network $\mathcal{G}$ that learns to align the latent distributions:
\begin{equation}
\tilde{\mathbf{z}}_{\text{LF}}(t) = \mathbf{z}_{\text{LF}}(t) + \mathcal{G}(\mathbf{z}_{\text{LF}}(t); \boldsymbol{\theta}_G).
\label{eq:gan_alignment}
\end{equation}
The generator $\mathcal{G}: \mathbb{R}^{d_z} \to \mathbb{R}^{d_z}$ is a shallow MLP with LeakyReLU activations, initialized to output near-zero corrections. A discriminator network $\mathcal{D}: \mathbb{R}^{d_z} \to [0,1]$ is trained adversarially \cite{GoodfellowGAN} to distinguish between latent codes from simulation and transformed codes from real data:
\begin{align}
\hspace*{-.1in} \mathcal{L}_D &\!=\! -\mathbb{E}_{\mathbf{z} \sim p_{\text{sim}}}\left[\log \mathcal{D}(\mathbf{z})\right] - \mathbb{E}_{\mathbf{z} \sim p_{\text{sim}}}\left[\log(1 - \mathcal{D}(\tilde{\mathbf{z}}))\right], \\
\hspace*{-.1in} \mathcal{L}_G &\!=\! -\mathbb{E}_{\mathbf{z} \sim p_{\text{sim}}}\left[\log \mathcal{D}(\tilde{\mathbf{z}})\right],
\label{eq:gan_loss}
\end{align}
where $\tilde{\mathbf{z}} = \mathbf{z} + \mathcal{G}(\mathbf{z})$ for $\mathbf{z}$ sampled from simulation latents, and the discriminator is trained to classify real sensor latents as real and generator-transformed simulation latents as fake.

\subsubsection{LF Decoder with Spectral Constraint}

The aligned latent code is decoded and then low-pass filtered to enforce the low-frequency constraint:
\begin{equation}
\tilde{\mathbf{u}}_{\text{LF}}(t) = \mathcal{P}_{k_c}\left(\mathcal{D}_{\text{LF}}(\tilde{\mathbf{z}}_{\text{LF}}(t); \boldsymbol{\theta}_{\text{dec}})\right) \in \mathbb{R}^n,
\label{eq:lf_decoder}
\end{equation}
where $\mathcal{D}_{\text{LF}}$ is a three-layer MLP with ReLU activations, and $\mathcal{P}_{k_c}$ denotes a low-pass filter that retains only Fourier modes with wavenumber $k \leq k_c$:
\begin{equation}
\mathcal{P}_{k_c}(\mathbf{u}) = \mathcal{F}^{-1}\left(\mathbf{1}_{k \leq k_c} \cdot \mathcal{F}(\mathbf{u})\right),
\label{eq:lowpass}
\end{equation}
where $\mathcal{F}$ and $\mathcal{F}^{-1}$ denote the discrete Fourier transform and its inverse. This explicit spectral constraint separates LF and HF components ~\cite{canuto2006spectral}.

\subsection{High-Frequency Pathway}

The HF pathway is designed to capture the fine-scale discrepancy between the LF reconstruction and reality. The input to the HF pathway is the sensor-space residual between observed measurements and LF predictions at sensor locations. Let $\mathbf{M} \in \mathbb{R}^{p \times n}$ denote the sensor sampling operator. The residual history is computed by subtracting the current LF prediction (at sensor locations) from each lag of the sensor history:
\begin{equation}
\mathbf{r}_{t} = \mathbf{s}'_{t} - \mathbf{M}\tilde{\mathbf{u}}_{\text{LF}}(t), \quad t = 0, 1, \ldots, m-1,
\label{eq:residual}
\end{equation}
yielding the residual time-history $\mathbf{R}_{t-L:t} = [\mathbf{r}_{t-L+1}, \ldots, \mathbf{r}_t] \in \mathbb{R}^{p \times L}$. Note that each lag uses its corresponding LF prediction $\tilde{\mathbf{u}}_{\text{LF}}(t-\ell)$, ensuring the residual captures the discrepancy between observed and predicted states at each time step.

The HF encoder employs an attention mechanism over temporal lags \cite{bahdanau2014neural} to learn which timesteps are most informative for predicting the HF correction, with time-derivative embedding to capture velocity and acceleration information (see Appendix~\ref{app:hf_details}). The HF decoder generates a base spatial pattern and applies a learned spatially-varying deformation to correct for velocity mismatches.

\subsection{Training Pipeline} 

The Cheap2Rich model is trained in four sequential stages to ensure stable learning of each component: (1) SHRED training on simulation data, (2) latent GAN training for distribution alignment, (3) HF-SHRED training with spectral sparsity regularization, and (4) fine-tuning. Complete training details including loss functions and hyperparameters are provided in Appendix~\ref{app:training}.

\subsection{Physical Interpretation}

The multi-scale decomposition has a natural physical interpretation in the context of RDE dynamics: \textbf{LF Component:} Captures the dominant detonation front dynamics that are well-represented by simplified models (e.g., Koch's model). The low-pass filter $\mathcal{P}_{k_c}$ ensures this component contains only large-scale spatial structures with wavenumber $k \leq k_c$. \textbf{HF Component:} Represents fine-scale corrections due to unmodeled injector dynamics, mixing processes, and turbulent fluctuations. The spectral sparsity regularization encourages the discovery of parsimonious, dominant correction modes, while the bandlimited penalty discourages energy at very high frequencies. \textbf{Time-Delay Embedding:} The temporal derivative augmentation and attention mechanism allow the HF pathway to capture velocity and phase information from the sensor history, enabling correction of wave propagation mismatches between simulation and reality.

\begin{figure}[t]
\centering
\includegraphics[width=\columnwidth]{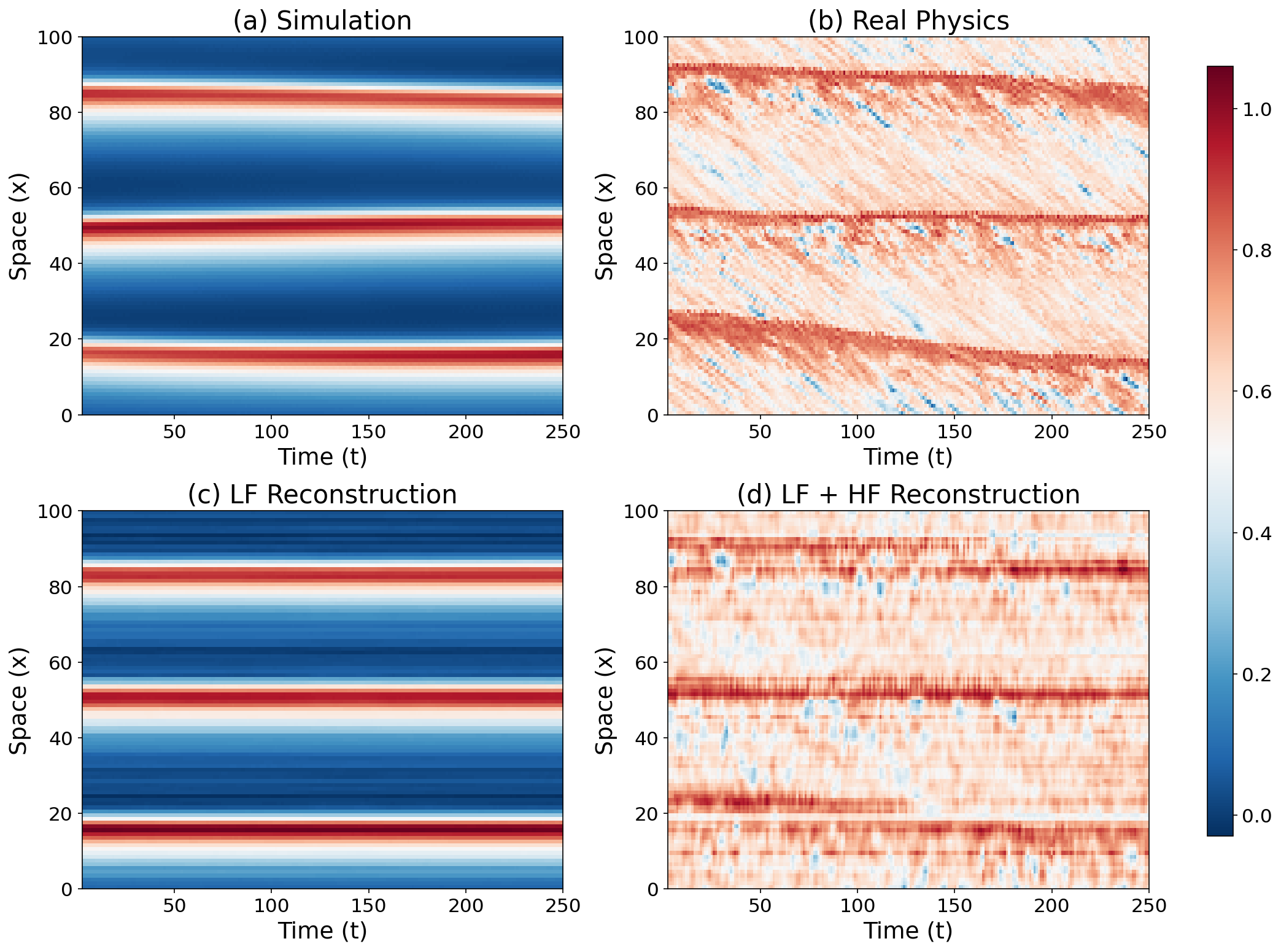}
\caption{Comparison on the full dataset. (a) Koch's model simulation capturing dominant front physics. (b) High-fidelity real physics with complex injector-driven dynamics. (c) LF reconstruction from the GAN-aligned SHRED pathway. (d) Full Cheap2Rich LF+HF reconstruction.}
\label{fig:four_panel}
\end{figure}

\section{Results}

We evaluate the Cheap2Rich framework on the rotating detonation engine dataset described in Section~2. The high-fidelity simulation data, which resolves the full three-dimensional reacting flow including shock dynamics, chemical kinetics, and injector interactions, requires weeks of supercomputer time to generate for a single operating condition. In contrast, Koch's one-dimensional model captures only the dominant detonation front propagation and can be computed in seconds. Our goal is to demonstrate that Cheap2Rich can bridge this fidelity gap using only sparse sensor measurements from the high-fidelity simulation, without requiring access to its full state.

\subsection{Experimental Setup}

The dataset consists of $m = 250$ temporal snapshots over a spatial domain discretized into $n = 100$ grid points. We use $p = 25$ uniformly distributed sensors and a temporal lag window of $L = 25$ time steps. The data is split into 80\% for training and 20\% for validation. The cutoff frequency for spectral constraints is set to $k_c = 12$. The training follows a three-stage pipeline: (1) SHRED pretraining on Koch's model simulation, (2) LF-DA-SHRED with latent GAN alignment, and (3) HF training with spectral sparsity ($\lambda = 0.1$, then fine-tuned with $\lambda' = 0.01$).

\subsection{Reconstruction Performance}

Figure~\ref{fig:four_panel} presents a qualitative comparison of the reconstruction pipeline. The Koch's model simulation (panel a) captures the three co-rotating detonation fronts but exhibits smooth, idealized dynamics. The high-fidelity simulation (panel b) reveals substantially richer structure: the detonation fronts display spatial variability, and significant fine-scale fluctuations arise from injector dynamics, mixing processes, and turbulent interactions that are absent in the simplified model.
The LF reconstruction (panel c) produces output that closely mirrors the Koch's model, as expected since the LF pathway is trained on simulation data. Despite the GAN-based latent alignment, the LF component alone cannot capture the fine-scale discrepancies, yielding an RMSE of 0.4114. The full Cheap2Rich LF+HF reconstruction (panel d) successfully recovers the complex dynamics of the high-fidelity simulation, reducing the RMSE to 0.1031---an improvement of 74.9\%.

\begin{table}[t]
\centering
\begin{tabular}{lcc}
\hline
\textbf{Method} & \textbf{RMSE (Valid)} & \textbf{SSIM} \\
\hline
Sim2Real Gap & 0.4110 & 0.1038 \\
LF-DA-SHRED & 0.4114 & 0.1113 \\
Cheap2Rich & \textbf{0.1031} & \textbf{0.3638} \\
\hline
Improvement & 74.9\% & 250.5\% \\
\hline
\end{tabular}
\caption{Reconstruction comparison. The Cheap2Rich framework achieves substantial improvement by learning spectrally-sparse corrections from sensor residuals - detailed comparison to other baseline approaches in \ref{app:baselines}.}
\label{tab:results}
\end{table}

\subsection{Quantitative Analysis}

Table~\ref{tab:results} summarizes the reconstruction performance. The baseline SHRED model trained on simulation and applied directly to real sensor data performs poorly due to the distribution shift. The GAN alignment provides minimal improvement for the LF pathway alone, indicating that the dominant discrepancy lies in unmodeled physics rather than latent distribution mismatch. The HF correction provides the critical improvement, reducing reconstruction error by over 74\% on the held-out validation set.

\subsection{Spectral Analysis of the HF Correction}

\begin{figure}[t]
\centering
\includegraphics[width=\columnwidth]{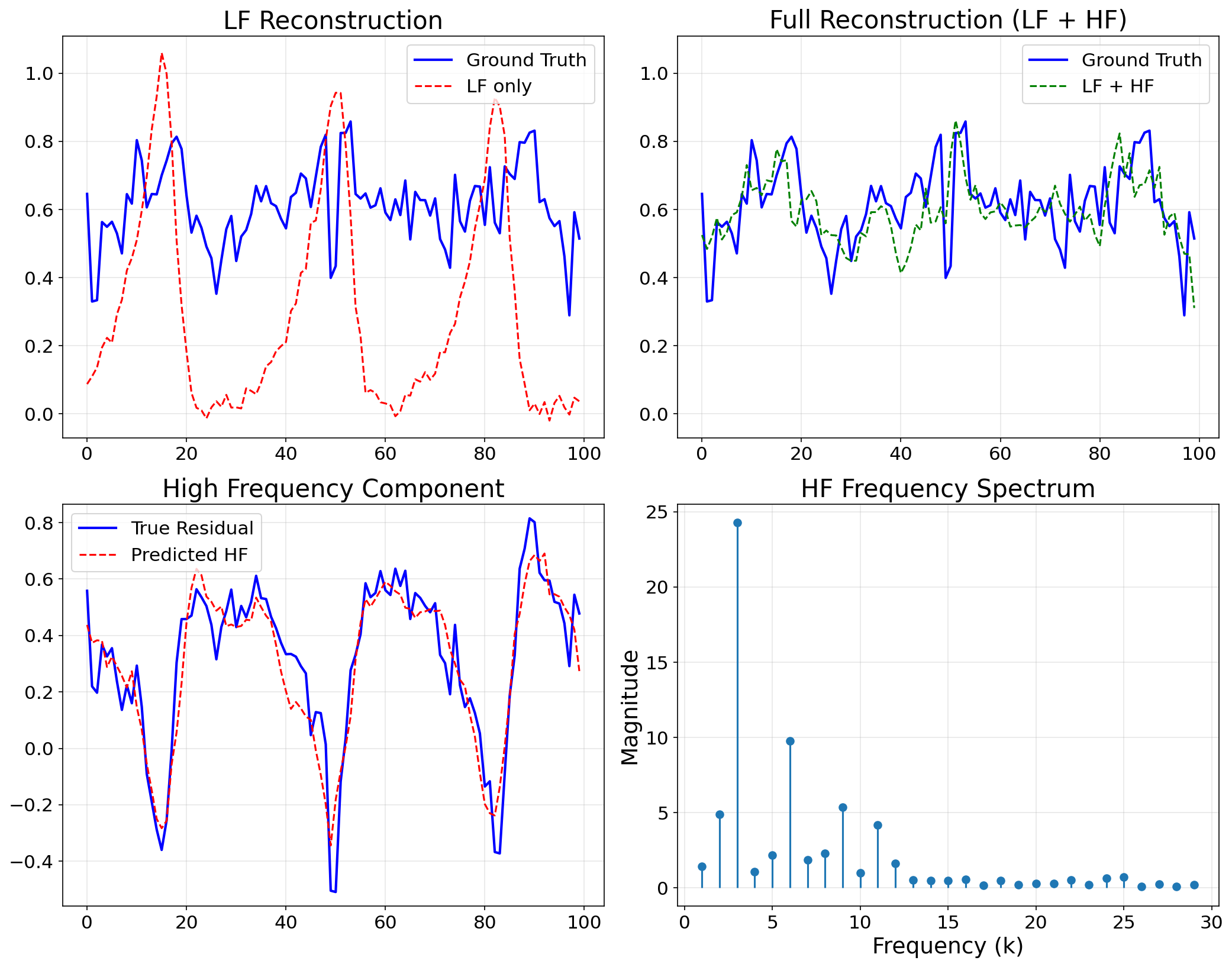}
\caption{Detailed analysis of multi-scale reconstruction. Top-left: LF reconstruction versus ground truth at a representative spatial slice. Top-right: Cheap2Rich LF+HF reconstruction. Bottom-left: Predicted HF component compared to the true residual. Bottom-right: Frequency spectrum of the HF correction showing sparse structure concentrated at wavenumbers $k \in \{3, 6, 9\}$.}
\label{fig:multiscale_results}
\vspace{-8mm}
\end{figure}

Figure~\ref{fig:multiscale_results} provides detailed analysis of the multi-scale decomposition. The top-left panel shows that the LF reconstruction fails to track the high-fidelity ground truth, while the top-right panel demonstrates that the full Cheap2Rich LF+HF reconstruction closely follows the true dynamics. The bottom-left panel compares the predicted HF component against the true residual (ground truth minus LF prediction), showing good agreement in both amplitude and phase.
The bottom-right panel reveals the spectral structure of the learned HF correction. The frequency content is sparse and concentrated at wavenumbers $k \in \{3, 6, 9, 11\}$, with the dominant modes at $k = 3, 6, 9$ corresponding to harmonics of the three co-rotating detonation waves. This physically interpretable structure emerges automatically from the spectral sparsity regularization, suggesting that the HF pathway has learned to represent the injector-driven physics in terms of modes that are commensurate with the underlying waves.

\subsection{Physical Interpretation}

The results demonstrate that the method successfully decomposes the Cheap2Rich gap into interpretable components:

\textbf{Front Physics (LF Component).} The LF pathway, trained on Koch's model data and aligned via the latent GAN, captures the dominant detonation front locations and propagation speeds. The three co-rotating fronts are clearly visible in both the simulation and LF reconstruction, confirming that the simplified model adequately represents the leading-order dynamics.

\textbf{Injector Physics (HF Component).} The HF correction captures the fine-scale dynamics that arise from physics absent in the Koch's model: injector response delays, spatially varying mixing efficiency \cite{schwer2010numerical}, and local turbulent fluctuations \cite{fotia2016experimental}. These effects manifest as structured corrections concentrated near the detonation fronts and in the inter-front regions where fresh reactants are injected.

The sparse spectral structure of the HF component---with energy concentrated at harmonics of the fundamental three-wave mode---suggests that the injector-driven physics are phase-locked to the detonation front passage. This is physically consistent with the periodic pressure fluctuations that modulate injector flow rates as each detonation wave passes.

\subsection{Computational Implications}

The Cheap2Rich framework enables high-fidelity state reconstruction at dramatically reduced computational cost. While the original high-fidelity simulation requires weeks of supercomputer time, our approach requires only sparse sensor measurements from the real system, a low-fidelity baseline model (e.g., Koch's model) that runs in seconds, and lightweight neural network inference that runs in minutes on a standard laptop without requiring GPU clusters.
This enables rapid exploration of the RDE design space, where each new operating condition can be characterized through sparse sensing rather than expensive full-scale simulation. The learned HF corrections provide interpretable diagnostics of injector performance without requiring direct measurement of the full injection dynamics.

\section{Discovery of Missing Physics via SINDy}

Beyond closing the simulation-to-reality gap through reconstruction, the Cheap2Rich framework enables explicit identification of the missing physics terms that account for the discrepancy between the simplified Koch's model and high-fidelity dynamics. Following the methodology established in \cite{bao2025data}, we employ sparse identification of nonlinear dynamics (SINDy) \cite{rudy2017data} within DA-SHRED architecture to discover interpretable governing equations for the learned corrections. The core insight is that the multi-scale decomposition provides natural targets for physics discovery: the LF correction dynamics, the HF component dynamics, and the direct missing physics (see Appendix~\ref{app:sindy} for the complete SINDy framework and library construction details).

\subsection{Discovered Equations}

Applying SINDy to the Cheap2Rich outputs yields the following discovered equations.

\textbf{Known Baseline: Koch's Model.} For reference, the simplified Koch's model governing the simulation takes the form:
{
\setlength{\abovedisplayskip}{4pt}
\setlength{\belowdisplayskip}{4pt}
\begin{equation}
\frac{\partial u}{\partial t} + u \frac{\partial u}{\partial x} = q \cdot k(1-\lambda) \exp\left(\frac{u - u_c}{\alpha}\right) - \epsilon u^2,
\label{eq:koch}
\end{equation}
}which captures the detonation front propagation through nonlinear advection and Arrhenius-type reaction kinetics.

\textbf{Discovered: LF Correction.} The GAN alignment contributes the following correction to the simulation dynamics:
{
\setlength{\abovedisplayskip}{4pt}
\setlength{\belowdisplayskip}{4pt}
\begin{align}
\frac{\partial}{\partial t}\left(u_{\text{LF}} - u_{\text{sim}}\right) = \; &{-}0.048 \, \Delta u + 0.005 \, u_{\text{LF}} - 0.013 \, u_{\text{LF}}^2 \nonumber \\
&{-} 0.056 \, u_{\text{LF}} \cdot \Delta u + \mathcal{O}(\partial_x).
\label{eq:discovered_lf}
\end{align}
}The dominant terms are polynomial in $u_{\text{LF}}$ and the correction $\Delta u = u_{\text{LF}} - u_{\text{sim}}$, suggesting that the GAN primarily adjusts the amplitude and baseline of the reconstruction rather than its spatial structure.

\textbf{Discovered: HF Dynamics.} The high-frequency component evolves according to:
{
\setlength{\abovedisplayskip}{4pt}
\setlength{\belowdisplayskip}{4pt}
\begin{align}
\label{eq:discovered_hf}
\frac{\partial u_{\text{HF}}}{\partial t} = \; &0.018 - 0.029 \, u_{\text{HF}} - 0.018 \, u_{\text{LF}}  \\
&{-} 0.067 \, u_{\text{LF}} \cdot u_{\text{HF}} + 0.151 \, u_{\text{LF}}^2 \cdot u_{\text{HF}} + \mathcal{O}(\partial_x). \nonumber
\end{align}
}The coupling terms $u_{\text{LF}} \cdot u_{\text{HF}}$ and $u_{\text{LF}}^2 \cdot u_{\text{HF}}$ indicate that the HF dynamics are modulated by the LF wave structure, consistent with the physical picture of injector-driven fluctuations being phase-locked to detonation front passage.

\textbf{Discovered: Direct Missing Physics.} The total discrepancy between real and simulated dynamics \eqref{eq:koch}  is governed by:
{
\setlength{\abovedisplayskip}{4pt}
\setlength{\belowdisplayskip}{4pt}
\begin{align}
\label{eq:discovered_missing}
\frac{\partial}{\partial t}\left(u_{\text{real}} - u_{\text{sim}}\right) = \; &0.723 - 3.79 \, u + 6.24 \, u^2 - 3.33 \, u^3 \nonumber \\ 
& \hspace*{-.5in} {+} 0.403 \, u_x - 0.404 \, u \cdot u_x + \mathcal{O}(u_{xx}).
\end{align}
}

\subsection{Physical Interpretation of Discovered Terms}

We discovered interpretable corrections to Koch's model, with each term addressing different physical mechanisms.

\textbf{Front Physics Corrections.} The direct missing physics equation \eqref{eq:discovered_missing} exhibits Burgers-type structure: a polynomial source term correcting reaction kinetics and advection terms adjusting wave speeds. The cubic polynomial $-3.79u + 6.24u^2 - 3.33u^3$ modifies the effective Arrhenius reaction rates, while $0.40 u_x - 0.40 u \cdot u_x$ corrects detonation front propagation. These terms primarily address deficiencies in Koch's representation of front physics.

\textbf{Injector Physics via LF-HF Coupling.} The HF dynamics equation \eqref{eq:discovered_hf} captures a different mechanism through the coupling terms $u_{\text{LF}} \cdot u_{\text{HF}}$ and $u_{\text{LF}}^2 \cdot u_{\text{HF}}$. This nonlinear modulation explains the observations from Section~4: HF corrections concentrate near detonation fronts and exhibit spectral sparsity at $k \in \{3, 6, 9\}$. The coupling ensures HF fluctuations are amplified in phase with the three-wave LF structure, consistent with injector-driven physics being modulated by periodic pressure fluctuations.

The discovered equations provide pathways for improving the Koch's model without requiring the neural network at inference time. Both direct correction using \eqref{eq:discovered_missing} and hierarchical correction using the LF and HF terms separately are possible; details are provided in Appendix~\ref{app:sindy}.

\section{Conclusions and Future Works}

This work presents a multi-scale data assimilation framework for rotating detonation engines that successfully bridges the sim2real gap between low-fidelity proxy models and complex coupled physics. The proposed Cheap2Rich architecture decomposes the reconstruction task into a low-frequency pathway, which captures dominant detonation front dynamics through GAN-aligned latent representations, and a high-frequency pathway that learns spectrally-sparse corrections from sensor residuals. Applied to a three-wave co-rotating RDE configuration, the framework achieves 80.9\% reduction in reconstruction RMSE while requiring only 25 sparse sensor measurements and lightweight neural network inference without GPU clusters. Spectral analysis reveals that the learned HF corrections exhibit sparse structure, with energy concentrated at harmonics of the fundamental three-wave mode, providing physically interpretable representations of injector-driven physics absent from the baseline proxy model. Furthermore, the application of SINDy to the multi-scale decomposition yields explicit governing equations for the missing physics, distinguishing between front physics corrections of Burgers type and injector physics captured through LF-HF coupling terms. These discovered functionals offer actionable pathways for improving reduced-order simulations.

Several promising directions emerge from this work. First, the trained Cheap2Rich model constitutes a computationally efficient surrogate that could enable gradient-based design optimization of RDE geometry, injector placement, and operating conditions at negligible computational cost compared to full-scale simulation. Such surrogate-driven optimization \cite{forrester2008engineering} would dramatically accelerate the exploration of high-performance engine configurations that currently require prohibitive computational resources. Second, Cheap2Rich's ability for full-state reconstruction from sparse real-time sensor measurements suggests potential applications in closed-loop control of RDE systems. By coupling the surrogate model with reinforcement learning algorithms \cite{rabault2019artificial}, one could develop controllers that modulate injector mass flow rates \cite{fotia2016experimental} to guarantee stable detonation with a prescribed number of waves \cite{bennewitz2018periodic}, addressing a critical challenge in transitioning RDE technology from laboratory demonstrations to operational propulsion systems. Third, while present work demonstrates Cheap2Rich on 1D RDE dynamics, extension to higher-dimensional spatiotemporal fields presents additional challenges that require substantial architectural innovations. Recent work has extended the LF/HF decomposition to 2D remote sensing applications using a hierarchical peeling structure, where the targets exhibit fundamentally different characteristics from the structured physics considered here: vegetation index reconstruction involves heterogeneous landscapes with sharp discontinuities, non-stationary spatial statistics, and irregular temporal dynamics without a physically well-established prior. Its success demonstrates that the Cheap2Rich paradigm has broader applicability beyond combustion physics and to higher dimensions---in turbulent flows, climate monitoring, and atmospheric modeling---where multi-fidelity approaches could dramatically reduce the computational burden of high-resolution simulation while preserving physically meaningful fine-scale structure. Fourth, the latent-space representations provide a natural foundation for autoregressive forecasting, wherein dynamics are evolved cheaply in the compressed space and subsequently decoded to full three-dimensional fields. Integrating SINDy-discovered dynamics into such a latent forecasting architecture \cite{champion2019data} could yield interpretable, long-horizon predictions that maintain physical consistency \cite{vlachas2022multiscale}. 




\section*{Impact Statement}

In this paper we investigate the applications of Machine Learning to the frontier of propulsion engineering - rotating detonation engines. The current bottlenecks in the design and engineering of RDEs stem from prohibitive simulation costs and our work demonstrates progress in finding inexpensive surrogate models which allow for design and control of these systems. Societal impacts from advancing propulsion technology are likely to increase humanity's footprint beyond Earth, but the same technologies can serve national security goals of various states.

\section*{Acknowledgments}
LLMs from OpenAI and Anthropic have been used in preparation of the code and corrections to the manuscript.
This work was supported in part by the US National Science Foundation (NSF) AI Institute for Dynamical Systems (dynamicsai.org), grant 2112085. JNK further acknowledges support from the Air Force Office of Scientific Research  (FA9550-24-1-0141).  The authors also acknowledge from the AFOSR/AFRL Center of Excellence in Assimilation of Flow Features in Compressible Reacting Flows under award number FA9550-25-1-0011, monitored by Dr. Chiping Li and Dr. Ramakanth Munipalli. MP acknowledges support from the National Defense Science and Engineering Graduate (NDSEG) Fellowship, USA through the Air Force Research Laboratory (AFRL).

\bibliography{refs}

@inproceedings{rahaman2019spectral,
  title={On the spectral bias of neural networks},
  author={Rahaman, Nasim and Baratin, Aristide and Arpit, Devansh and Draxler, Felix and Lin, Min and Hamprecht, Fred and Bengio, Yoshua and Courville, Aaron},
  booktitle={International conference on machine learning},
  pages={5301--5310},
  year={2019},
  organization={PMLR}
}

@article{wyder2025common,
  title={Common task framework for a critical evaluation of scientific machine learning algorithms},
  author={Wyder, Philippe Martin and Goldfeder, Judah and Yermakov, Alexey and Zhao, Yue and Riva, Stefano and Williams, Jan P and Zoro, David and Rude, Amy Sara and Tomasetto, Matteo and Germany, Joe and others},
  journal={arXiv preprint arXiv:2510.23166},
  year={2025}
}

@article{tomasetto2025reduced,
  title={Reduced order modeling with shallow recurrent decoder networks},
  author={Tomasetto, Matteo and Williams, Jan P and Braghin, Francesco and Manzoni, Andrea and Kutz, J Nathan},
  journal={Nature Communications},
  volume={16},
  number={1},
  pages={10260},
  year={2025},
  publisher={Nature Publishing Group UK London}
}

@book{brunton2022data,
  title={Data-driven science and engineering: Machine learning, dynamical systems, and control},
  author={Brunton, Steven L and Kutz, J Nathan},
  year={2022},
  publisher={Cambridge University Press}
}

@article{karniadakis2021physics,
  title={Physics-informed machine learning},
  author={Karniadakis, George Em and Kevrekidis, Ioannis G and Lu, Lu and Perdikaris, Paris and Wang, Sifan and Yang, Liu},
  journal={Nature Reviews Physics},
  volume={3},
  number={6},
  pages={422--440},
  year={2021},
  publisher={Nature Publishing Group UK London}
}

@inbook{Megan_RDE,
author = {Megan Powers and Shivank Sharma and Venkatramanan Raman},
title = {Impact of Varying Combustion Chamber Area in Rotating Detonation Engines},
booktitle = {AIAA SCITECH 2026 Forum},
year = {2026},
doi = {10.2514/6.2026-2613},
URL = {https://arc.aiaa.org/doi/abs/10.2514/6.2026-2613},
eprint = {https://arc.aiaa.org/doi/pdf/10.2514/6.2026-2613}
}

@article{KochRDE,
    author = {Koch, James and Kutz, J. Nathan},
    title = {Modeling thermodynamic trends of rotating detonation engines},
    journal = {Physics of Fluids},
    volume = {32},
    number = {12},
    pages = {126102},
    year = {2020},
    month = {12},
    issn = {1070-6631},
    doi = {10.1063/5.0023972},
    url = {https://doi.org/10.1063/5.0023972},
    eprint = {https://pubs.aip.org/aip/pof/article-pdf/doi/10.1063/5.0023972/12654141/126102_1_online.pdf},
}

@article{KochDD_RDE,
  title = {Data-driven modeling of rotating detonation waves},
  author = {Mendible, Ariana and Koch, James and Lange, Henning and Brunton, Steven L. and Kutz, J. Nathan},
  journal = {Phys. Rev. Fluids},
  volume = {6},
  issue = {5},
  pages = {050507},
  numpages = {20},
  year = {2021},
  month = {May},
  publisher = {American Physical Society},
  doi = {10.1103/PhysRevFluids.6.050507},
  url = {https://link.aps.org/doi/10.1103/PhysRevFluids.6.050507}
}

@article{Koch_multiscale_Physcics,
  title = {Multiscale physics of rotating detonation waves: Autosolitons and modulational instabilities},
  author = {Koch, James and Kurosaka, Mitsuru and Knowlen, Carl and Kutz, J. Nathan},
  journal = {Phys. Rev. E},
  volume = {104},
  issue = {2},
  pages = {024210},
  numpages = {17},
  year = {2021},
  month = {Aug},
  publisher = {American Physical Society},
  doi = {10.1103/PhysRevE.104.024210},
  url = {https://link.aps.org/doi/10.1103/PhysRevE.104.024210}
}

@article{VenkatNonlinearities_RDE,
   author = "Raman, Venkat and Prakash, Supraj and Gamba, Mirko",
   title = "Nonidealities in Rotating Detonation Engines", 
   journal= "Annual Review of Fluid Mechanics",
   year = "2023",
   volume = "55",
   number = "Volume 55, 2023",
   pages = "639-674",
   doi = "https://doi.org/10.1146/annurev-fluid-120720-032612",
   url = "https://www.annualreviews.org/content/journals/10.1146/annurev-fluid-120720-032612",
   publisher = "Annual Reviews",
   issn = "1545-4479",
   type = "Journal Article",
  }

@article{Mixing_and_detonation_RDE,
title = {Mixing and detonation structure in a rotating detonation engine with an axial air inlet},
journal = {Proceedings of the Combustion Institute},
volume = {38},
number = {3},
pages = {3769-3776},
year = {2021},
issn = {1540-7489},
doi = {https://doi.org/10.1016/j.proci.2020.06.283},
url = {https://www.sciencedirect.com/science/article/pii/S1540748920303758},
author = {Takuma Sato and Fabian Chacon and Logan White and Venkat Raman and Mirko Gamba}
}

@article{Experimental_Visualization_of_RDEs,
author = {Nakagami, Soma and Matsuoka, Ken and Kasahara, Jiro and Kumazawa, Yoshiki and Fujii, Jumpei and Matsuo, Akiko and Funaki, Ikkoh},
title = {Experimental Visualization of the Structure of Rotating Detonation Waves in a Disk-Shaped Combustor},
journal = {Journal of Propulsion and Power},
volume = {33},
number = {1},
pages = {80-88},
year = {2017},
doi = {10.2514/1.B36084},
URL = {         https://doi.org/10.2514/1.B36084},
eprint = {         https://doi.org/10.2514/1.B36084}
}

@article{Space_Flight_of_RDE,
author = {Sato, Tomoki and Matsuoka, Ken and Itouyama, Noboru and Yasui, Masaaki and Matsuyama, Koichi and Ide, Yuichiro and Nakata, Kotaro and Suzuki, Yamato and Ishibashi, Ryoto and Suzuki, Sota and Kasahara, Jiro and Kawasaki, Akira and Hirashima, Hidetoshi and Nakata, Daisuke and Eguchi, Hikaru and Takano, Tomoyuki and Uchiumi, Masaharu and Himeno, Takehiro and Yahata, Yusei and Yamada, Kazuhiko},
year = {2025},
month = {11},
pages = {1-18},
title = {Space Flight of Liquid Rotating Detonation Engine Using Sounding Rocket S-520-34},
journal = {Journal of Spacecraft and Rockets},
doi = {10.2514/1.A36447}
}

@article{bao2025data,
  title={Data assimilation and discrepancy modeling with shallow recurrent decoders},
  author={Bao, Yuxuan and Kutz, J Nathan},
  journal={arXiv preprint arXiv:2512.01170},
  year={2025}
}

@article{brunton2016discovering,
  title={Discovering governing equations from data by sparse identification of nonlinear dynamical systems},
  author={Brunton, Steven L and Proctor, Joshua L and Kutz, J Nathan},
  journal={Proceedings of the national academy of sciences},
  volume={113},
  number={15},
  pages={3932--3937},
  year={2016},
  publisher={National Academy of Sciences}
}

@incollection{DA_ocean,
title = {Data Assimilation in Meteorology and Oceanography},
editor = {Renata Dmowska and Barry Saltzman},
series = {Advances in Geophysics},
publisher = {Elsevier},
volume = {33},
pages = {141-266},
year = {1991},
issn = {0065-2687},
doi = {https://doi.org/10.1016/S0065-2687(08)60442-2},
url = {https://www.sciencedirect.com/science/article/pii/S0065268708604422},
author = {Michael Ghil and Paola Malanotte-Rizzoli}
}

@Article{da_review,
AUTHOR = {Bocquet, M. and Brajard, J. and Carrassi, A. and Bertino, L.},
TITLE = {Data assimilation as a learning tool to infer ordinary differential equation representations of dynamical models},
JOURNAL = {Nonlinear Processes in Geophysics},
VOLUME = {26},
YEAR = {2019},
NUMBER = {3},
PAGES = {143--162},
URL = {https://npg.copernicus.org/articles/26/143/2019/},
DOI = {10.5194/npg-26-143-2019}
}

@article{sensing_with_shred,
    author = {Williams, Jan P. and Zahn, Olivia and Kutz, J. Nathan},
    title = {Sensing with shallow recurrent decoder networks},
    journal = {Proceedings of the Royal Society A: Mathematical, Physical and Engineering Sciences},
    volume = {480},
    number = {2298},
    pages = {20240054},
    year = {2024},
    month = {09},
    issn = {1364-5021},
    doi = {10.1098/rspa.2024.0054},
    url = {https://doi.org/10.1098/rspa.2024.0054},
    eprint = {https://royalsocietypublishing.org/rspa/article-pdf/doi/10.1098/rspa.2024.0054/513220/rspa.2024.0054.pdf},
}

@article{GPU-solver-for-RDEs,
author = {Carreon, Anthony and Singh, Jagmohan and Sharma, Shivank and Zhang, Shuzhi and Raman, Venkat},
year = {2025},
month = {10},
pages = {},
journal = {},
title = {A GPU-based Compressible Combustion Solver for Applications Exhibiting Disparate Space and Time Scales},
doi = {10.48550/arXiv.2510.23993}
}

@article{AMReX,
author = {Zhang, Weiqun and Almgren, Ann and Beckner, Vince and Bell, John and Blaschke, Johannes and Chan, Cy and Day, Marcus and Friesen, Brian and Gott, Kevin and Graves, Daniel and Katz, Max and Myers, Andrew and Nguyen, Tan and Nonaka, Andrew and Rosso, Michele and Williams, Samuel and Zingale, Michael},
year = {2019},
month = {05},
pages = {1370},
title = {AMReX: a framework for block-structured adaptive mesh refinement},
volume = {4},
journal = {Journal of Open Source Software},
doi = {10.21105/joss.01370}
}

@article{FoundationalFuelChem,
  title={Foundational fuel chemistry model version 1.0 (FFCM-1)},
  author={Smith, Gregory P and Tao, Y and Wang, H},
  journal={epub, accessed July},
  volume={26},
  pages={2018},
  year={2016}
}

@misc{FFCMY_personal,
  author        = {Xu, R. and Wang, H.},
  title         = {A Reduced Reaction Model of Methane Combustion: FFCMy-12},
  howpublished  = {Personal communication},
  year          = {2018}
}

@article{pyclaw-sisc,
        Author = {Ketcheson, David I. and Mandli, Kyle T. and Ahmadia, Aron J. and Alghamdi, Amal and {Quezada de Luna}, Manuel and Parsani, Matteo and Knepley, Matthew G. and Emmett, Matthew},
        Journal = {SIAM Journal on Scientific Computing},
        Month = nov,
        Number = {4},
        Pages = {C210--C231},
        Title = {{PyClaw: Accessible, Extensible, Scalable Tools for Wave Propagation Problems}},
        Volume = {34},
        Year = {2012}}

@inproceedings{GoodfellowGAN,
author = {Goodfellow, Ian J. and Pouget-Abadie, Jean and Mirza, Mehdi and Xu, Bing and Warde-Farley, David and Ozair, Sherjil and Courville, Aaron and Bengio, Yoshua},
title = {Generative adversarial nets},
year = {2014},
publisher = {MIT Press},
address = {Cambridge, MA, USA},
booktitle = {Proceedings of the 28th International Conference on Neural Information Processing Systems - Volume 2},
pages = {2672–2680},
numpages = {9},
location = {Montreal, Canada},
series = {NIPS'14}
}

@article{hochreiter1997long,
  title={Long short-term memory},
  author={Hochreiter, Sepp and Schmidhuber, J{\"u}rgen},
  journal={Neural computation},
  volume={9},
  number={8},
  pages={1735--1780},
  year={1997},
  publisher={MIT press}
}

@book{canuto2006spectral,
  title={Spectral methods: fundamentals in single domains},
  author={Canuto, Claudio and Hussaini, M Youssuff and Quarteroni, Alfio and Zang, Thomas A},
  year={2006},
  publisher={Springer}
}

@article{bahdanau2014neural,
  title={Neural machine translation by jointly learning to align and translate},
  author={Bahdanau, Dzmitry and Cho, Kyunghyun and Bengio, Yoshua},
  journal={arXiv preprint arXiv:1409.0473},
  year={2014}
}

@inproceedings{schwer2010numerical,
  title={Numerical investigation of rotating detonation engines},
  author={Schwer, Douglas and Kailasanath, Kailas},
  booktitle={46th AIAA/ASME/SAE/ASEE joint propulsion conference \& exhibit},
  pages={6880},
  year={2010}
}

@article{fotia2016experimental,
  title={Experimental study of the performance of a rotating detonation engine with nozzle},
  author={Fotia, Matthew L and Schauer, Fred and Kaemming, Tom and Hoke, John},
  journal={Journal of Propulsion and Power},
  volume={32},
  number={3},
  pages={674--681},
  year={2016},
  publisher={American Institute of Aeronautics and Astronautics}
}

@article{rudy2017data,
  title={Data-driven discovery of partial differential equations},
  author={Rudy, Samuel H and Brunton, Steven L and Proctor, Joshua L and Kutz, J Nathan},
  journal={Science advances},
  volume={3},
  number={4},
  pages={e1602614},
  year={2017},
  publisher={American Association for the Advancement of Science}
}

@article{ba2016layer,
  title={Layer normalization},
  author={Ba, Jimmy Lei and Kiros, Jamie Ryan and Hinton, Geoffrey E},
  journal={arXiv preprint arXiv:1607.06450},
  year={2016}
}

@book{forrester2008engineering,
  title={Engineering design via surrogate modelling: a practical guide},
  author={Forrester, Alexander and Sobester, Andras and Keane, Andy},
  year={2008},
  publisher={John Wiley \& Sons}
}

@article{champion2019data,
  title={Data-driven discovery of coordinates and governing equations},
  author={Champion, Kathleen and Lusch, Bethany and Kutz, J Nathan and Brunton, Steven L},
  journal={Proceedings of the National Academy of Sciences},
  volume={116},
  number={45},
  pages={22445--22451},
  year={2019},
  publisher={National Academy of Sciences}
}

@article{vlachas2022multiscale,
  title={Multiscale simulations of complex systems by learning their effective dynamics},
  author={Vlachas, Pantelis R and Arampatzis, Georgios and Uhler, Caroline and Koumoutsakos, Petros},
  journal={Nature Machine Intelligence},
  volume={4},
  number={4},
  pages={359--366},
  year={2022},
  publisher={Nature Publishing Group UK London}
}

@article{bennewitz2018periodic,
  title={Periodic partial extinction in acoustically coupled fuel droplet combustion},
  author={Bennewitz, John W and Valentini, Dario and Plascencia, Miguel A and Vargas, Andres and Sim, Hyung Sub and Lopez, Brett and Smith, Owen I and Karagozian, Ann R},
  journal={Combustion and flame},
  volume={189},
  pages={46--61},
  year={2018},
  publisher={Elsevier}
}

@article{rabault2019artificial,
  title={Artificial neural networks trained through deep reinforcement learning discover control strategies for active flow control},
  author={Rabault, Jean and Kuchta, Miroslav and Jensen, Atle and R{\'e}glade, Ulysse and Cerardi, Nicolas},
  journal={Journal of fluid mechanics},
  volume={865},
  pages={281--302},
  year={2019},
  publisher={Cambridge University Press}
}

@article{Detonative_propulsion_review,
title = {Detonative propulsion},
journal = {Proceedings of the Combustion Institute},
volume = {34},
number = {1},
pages = {125-158},
year = {2013},
issn = {1540-7489},
doi = {https://doi.org/10.1016/j.proci.2012.10.005},
url = {https://www.sciencedirect.com/science/article/pii/S1540748912004014},
author = {Piotr Wolański}
}

@article{bennewitz2019modal,
  title={Modal transitions in rotating detonation rocket engines},
  author={Bennewitz, John W and Bigler, Blaine R and Pilgram, Jessica J and Hargus Jr, William A},
  journal={International Journal of Energetic Materials and Chemical Propulsion},
  volume={18},
  number={2},
  year={2019},
  publisher={Begel House Inc.}
}

@article{prakash2021numerical,
  title={Numerical simulation of a methane-oxygen rotating detonation rocket engine},
  author={Prakash, Supraj and Raman, Venkat and Lietz, Christopher F and Hargus Jr, William A and Schumaker, Stephen A},
  journal={Proceedings of the Combustion Institute},
  volume={38},
  number={3},
  pages={3777--3786},
  year={2021},
  publisher={Elsevier}
}

@article{numericsHMM,
      title={An AMReX-based compressible reacting flow solver for high-speed reacting flows relevant to hypersonic propulsion},
      author={Shivank Sharma and Ral Bielawski and Oliver Gibson and Shuzhi Zhang and Vansh Sharma and Andreas H. Rauch and Jagmohan Singh and Sebastian Abisleiman and Michael Ullman and Shivam Barwey and Venkat Raman},
      year={2024},
      journal={arXiv preprint arXiv:2412.00900}
}

@article{bielawski2023highly,
  title={Highly-scalable GPU-accelerated compressible reacting flow solver for modeling high-speed flows},
  author={Bielawski, Ral and Barwey, Shivam and Prakash, Supraj and Raman, Venkat},
  journal={Computers \& Fluids},
  volume={265},
  pages={105972},
  year={2023},
  publisher={Elsevier}
}

@article{roy2025physics,
  title={A Physics-informed Multi-resolution Neural Operator},
  author={Roy, Sumanta and Bahmani, Bahador and Kevrekidis, Ioannis G and Shields, Michael D},
  journal={arXiv preprint arXiv:2510.23810},
  year={2025}
}

@article{li2020fourier,
  title={Fourier neural operator for parametric partial differential equations},
  author={Li, Zongyi and Kovachki, Nikola and Azizzadenesheli, Kamyar and Liu, Burigede and Bhattacharya, Kaushik and Stuart, Andrew and Anandkumar, Anima},
  journal={arXiv preprint arXiv:2010.08895},
  year={2020}
}

@article{lu2021learning,
  title={Learning nonlinear operators via DeepONet based on the universal approximation theorem of operators},
  author={Lu, Lu and Jin, Pengzhan and Pang, Guofei and Zhang, Zhongqiang and Karniadakis, George Em},
  journal={Nature machine intelligence},
  volume={3},
  number={3},
  pages={218--229},
  year={2021},
  publisher={Nature Publishing Group UK London}
}

@article{kutz2025accelerating,
  title={Accelerating scientific discovery with the common task framework},
  author={Kutz, J Nathan and Battaglia, Peter and Brenner, Michael and Carlberg, Kevin and Hagberg, Aric and Ho, Shirley and Hoyer, Stephan and Lange, Henning and Lipson, Hod and Mahoney, Michael W and others},
  journal={arXiv preprint arXiv:2511.04001},
  year={2025}
}

@article{erichson2020shallow,
  title={Shallow neural networks for fluid flow reconstruction with limited sensors},
  author={Erichson, N Benjamin and Mathelin, Lionel and Yao, Zhewei and Brunton, Steven L and Mahoney, Michael W and Kutz, J Nathan},
  journal={Proceedings of the Royal Society A},
  volume={476},
  number={2238},
  pages={20200097},
  year={2020},
  publisher={The Royal Society Publishing}
}

@article{jiang2025hierarchical,
  title={Hierarchical Implicit Neural Emulators},
  author={Jiang, Ruoxi and Zhang, Xiao and Jakhar, Karan and Lu, Peter Y and Hassanzadeh, Pedram and Maire, Michael and Willett, Rebecca},
  journal={arXiv preprint arXiv:2506.04528},
  year={2025}
}

@article{fan2025physics,
  title={Physics-Informed Inference Time Scaling via Simulation-Calibrated Scientific Machine Learning},
  author={Fan, Zexi and Sun, Yan and Yang, Shihao and Lu, Yiping},
  journal={arXiv preprint arXiv:2504.16172},
  year={2025}
}

@article{niu2024multi,
  title={Multi-fidelity residual neural processes for scalable surrogate modeling},
  author={Niu, Ruijia and Wu, Dongxia and Kim, Kai and Ma, Yi-An and Watson-Parris, Duncan and Yu, Rose},
  journal={arXiv preprint arXiv:2402.18846},
  year={2024}
}

@article{liu2024kan,
  title={Kan: Kolmogorov-arnold networks},
  author={Liu, Ziming and Wang, Yixuan and Vaidya, Sachin and Ruehle, Fabian and Halverson, James and Solja{\v{c}}i{\'c}, Marin and Hou, Thomas Y and Tegmark, Max},
  journal={arXiv preprint arXiv:2404.19756},
  year={2024}
}
\bibliographystyle{icml2026}

\newpage
\appendix
\onecolumn
\section{Preprocessing of the High Fidelity Sim}
\label{app:preprocessing}
\subsection{Interpolation of ADR onto a fixed spatial grid}

To map each snapshot from the native (unstructured) simulation point cloud onto a fixed cylindrical reference grid with $3 \times 100 \times 100$ points equally spaced in the $ r, \phi , z$ directions in a cylindrical coordinate system, we build a KD-tree over the source coordinates $\{\mathbf{x}_j\}_{j=1}^{N_s}\subset\mathbb{R}^3$ and, for every target grid point $\mathbf{g}_i$ ($i=1,\dots,N_g$), query its $k$ nearest neighbors $\mathcal{N}_k(i)$ with Euclidean distances $d_{ij}=\|\mathbf{g}_i-\mathbf{x}_j\|_2$. The field value $v(\mathbf{g}_i)$ is then obtained by inverse-distance weighting with a softened exponent, i.e.,
\[
\hat v(\mathbf{g}_i)=\sum_{j\in\mathcal{N}_k(i)} w_{ij}\, v(\mathbf{x}_j),\quad
w_{ij}=\frac{\left(d_{ij}+\varepsilon\right)^{-1/2}}{\sum_{\ell\in\mathcal{N}_k(i)} \left(d_{i\ell}+\varepsilon\right)^{-1/2}},
\]
where $\varepsilon$ is a small constant for numerical stability; this yields a smooth local interpolation while remaining computationally efficient via batched neighbor queries.

\subsection{Projection onto a 1d ring}

Finally, we pick $h_* = 20\,\mathrm{mm}$ and take the mean of the three points of the grid at the fixed $h_*$ and $ \phi(i) = \frac{i }{100} 2\pi$ obtaining a 1d dataset which is presented on Figure~\ref{fig:two-panels}. After the projection step, we also counter the rotation of the waves by shifting the frame at time index $i$ by $ i \frac{2 \pi}{250}$ radians, and changing the Temperature scale with a min-max scaler. 

\section{Koch's Model}
\label{app:KochModel}

The physical system describes a compressible reactive flow where fuel and oxidizer are continuously injected into an annular combustor channel, mix at a rate governed by a mixing parameter $s$, and undergo chemical reaction according to Arrhenius kinetics with Damköhler number $Da$ and activation energy $Ea$. The system conserves mass, momentum, energy, and a progress variable (mixture fraction $z$) that tracks the degree of mixing and reaction. Key physical phenomena include injection through the boundary characterized by an area ratio $AR$, heat release $hv$ from chemical reactions controlled by an ignition temperature $T_\text{ign}$, and a choked flow condition at the injection boundary that depends on the local pressure state. The governing equations employ a gamma-law equation of state with specific heat ratio $\gamma = 1.29$, representative of detonation products, and include Heaviside functions to model activation of injection, chemical reactions, and boundary conditions.

The numerical solution employs a finite volume method with operator splitting to separately handle hyperbolic transport and stiff reactive source terms. The inviscid flux terms are discretized using a second-order Clawpack framework \cite{pyclaw-sisc} with an HLLC (Harten-Lax-van Leer-Contact) Riemann solver that resolves shock waves, contact discontinuities, and expansion fans in the reactive flow field. The source terms arising from injection, mixing, and chemical reaction are integrated using a second-order explicit Runge-Kutta method (RK2) with adaptive time stepping controlled by a CFL condition of 0.1. The chemical source term incorporates a temperature-dependent reaction rate with Arrhenius kinetics, activated only above a threshold temperature of 1.01×Tign to ensure numerical stability. Periodic boundary conditions are enforced at both domain boundaries to simulate the azimuthal periodicity of the annular combustor, and the simulation tracks primitive variables (density, velocity, pressure, temperature, and mixture fraction) over a domain length of 24 characteristic lengths discretized with 100 grid points, evolving the solution to a final dimensionless time of 180 to capture multiple detonation wave passages and establish quasi-steady periodic behavior.

\begin{figure}[t]
    \centering

    \begin{subfigure}[b]{0.48\textwidth}
        \centering
        \includegraphics[width= \textwidth]{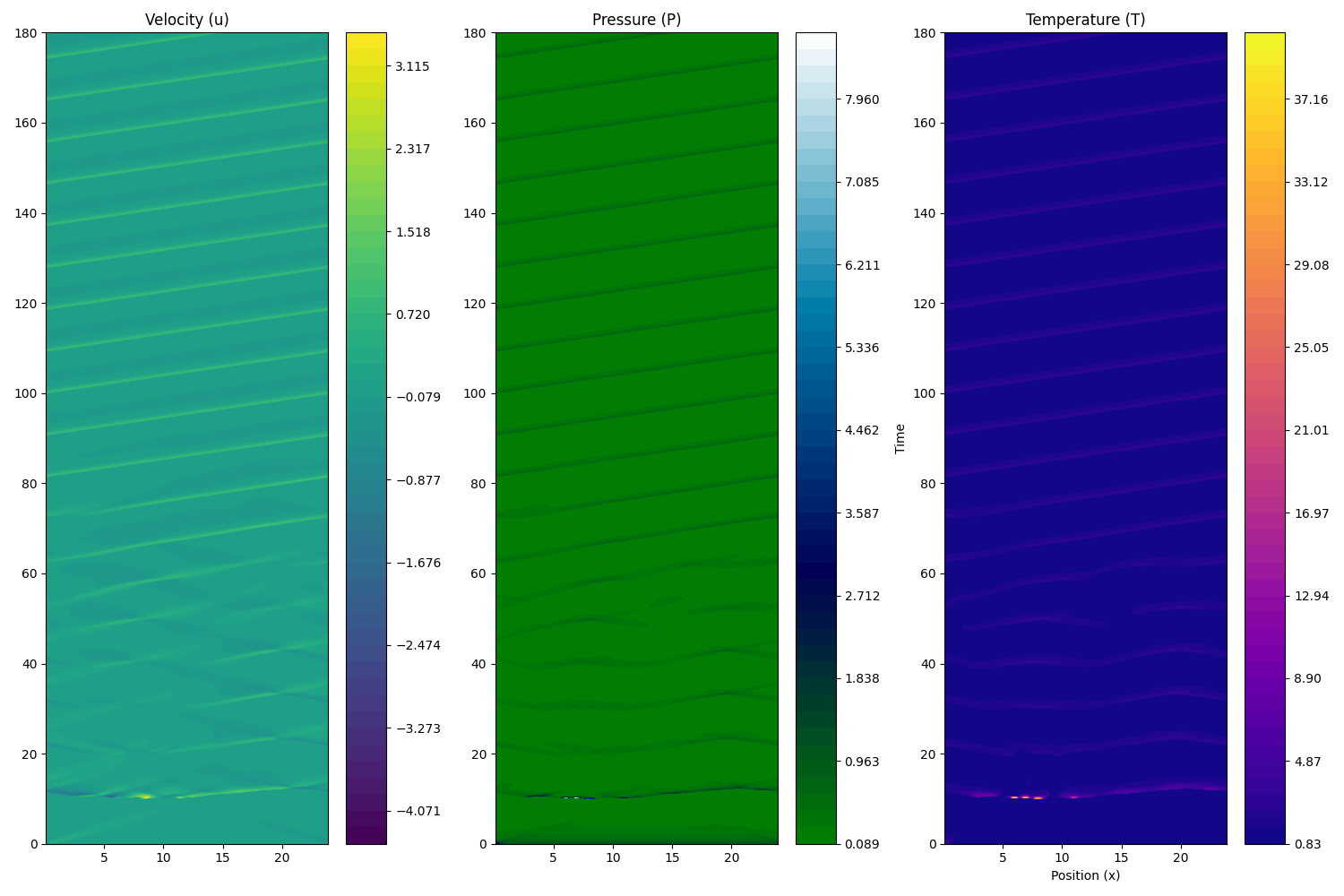}
        \caption{One Wave}
        \label{fig:one_wave}
    \end{subfigure}\hfill
    \begin{subfigure}[b]{0.48\textwidth}
        \centering
        \includegraphics[width=\textwidth]{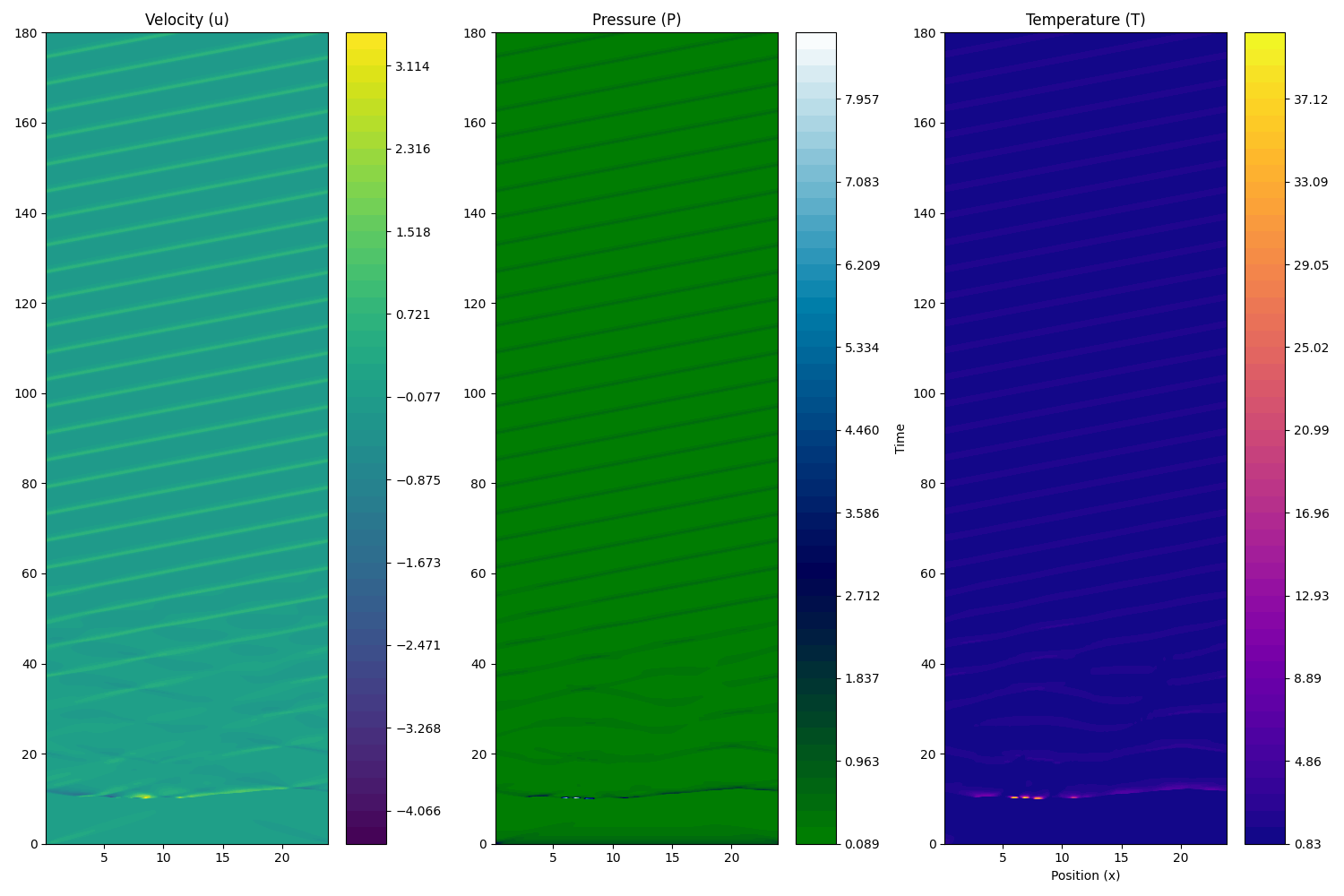}
        \caption{Two Waves}
        \label{fig:two_waves}
    \end{subfigure}

    \vspace{0.5em}

    \begin{subfigure}[b]{0.48\textwidth}
        \centering
        \includegraphics[width=\textwidth]{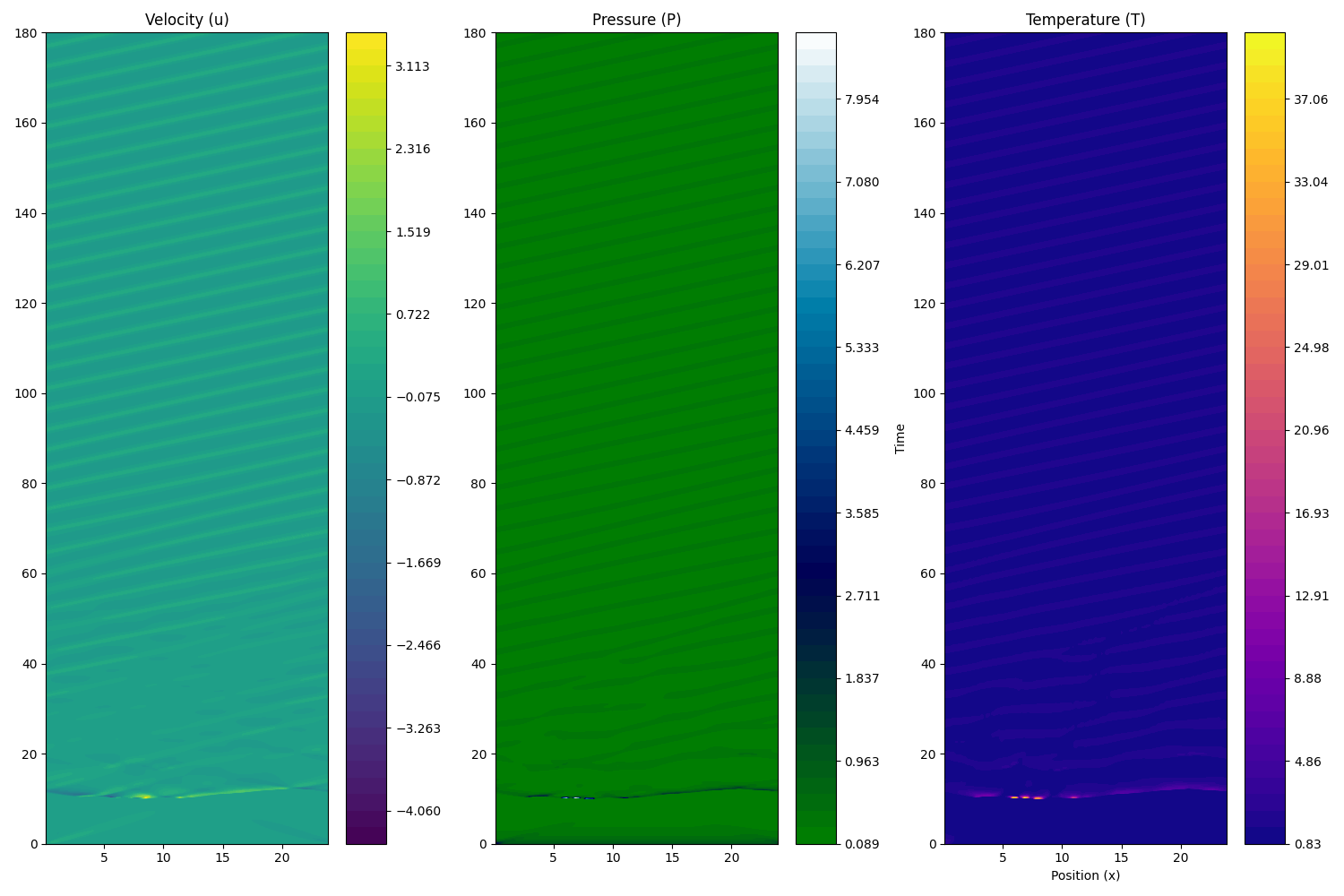}
        \caption{Three Waves}
        \label{fig:three_waves}
    \end{subfigure}\hfill
    \begin{subfigure}[b]{0.48\textwidth}
        \centering
        \includegraphics[width=\textwidth]{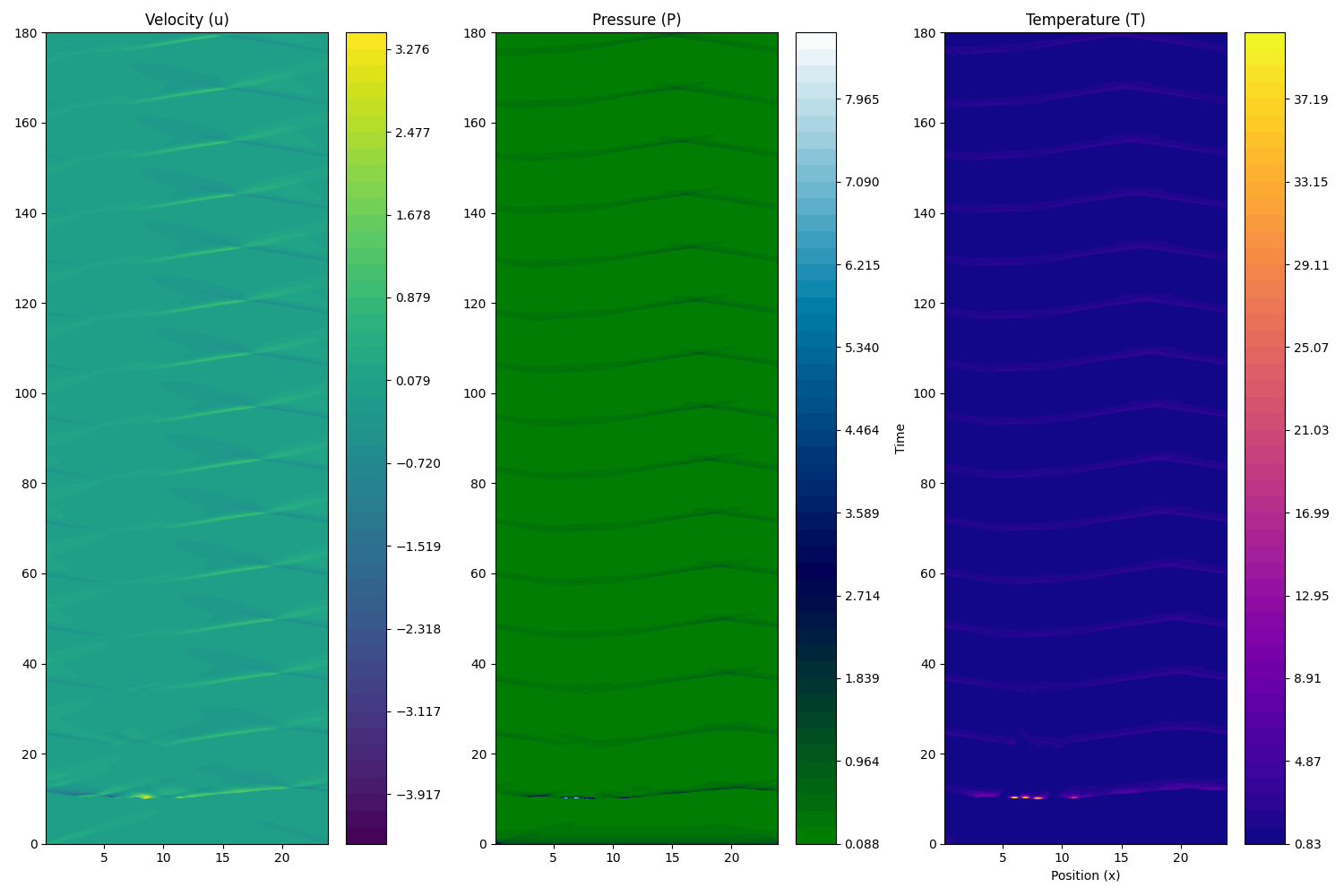}
        \caption{Pulsing detonation}
        \label{fig:pulsing}
    \end{subfigure}

    \caption{Comparison of four RDE operational modes demonstrating model capacity to simulate different wave structures}
    \label{fig:rde_comparison}
\end{figure}

Four distinct RDE configurations were investigated to demonstrate the universality of the model \ref{fig:rde_comparison}. Just by changing the $s$ parameter in the model we can obtain simulations which have quasi-steady states of one ($s = 0.05$) or two waves ($s = 0.06$), a pulsing detonation ($s = 0.04$), and three waves which correspond to the high-fidelity setting with ($s = 0.07$).

We find a set of parameters which best resemble the data observed in the high resolution 3d simulation. For the following set of parameters we are able to achieve a quasi-steady state of three co-rotating waves \ref{tab:assim_params}. 

\begin{table}[t]
    \centering
    \begin{tabular}{ll ll}
        \toprule
        \textbf{Parameter} & \textbf{Value} & \textbf{Parameter} & \textbf{Value} \\
        \midrule
        $\gamma$         & 1.29   & $Da$              & 289   \\
        $p_{\text{ref}}$ & 1.0    & $T_{\text{ign}}$  & 5.8   \\
        $\rho_{\text{ref}}$ & 1.0 & $h_v$            & 24.6  \\
        $R$              & 1.0    & $E_a$             & 11.5  \\
        $T_{\text{ref}}$ & 1.0    & $L$               & 24.0  \\
        $AR$             & 0.2    & $m_x$             & 100   \\
        $s$              & 0.07   & $t_{\text{final}}$& 180.0 \\
        $\beta$          & 14.286 &                  &       \\
        \bottomrule
    \end{tabular}
    \caption{Simulation parameters for three co-rotating waves configuration.}
    \label{tab:assim_params}
\end{table}

\newpage
\section{Notation Summary}
\label{app:notation}

See Table~\ref{tab:notation}.

\begin{table}[t]
\centering
\small
\begin{tabular}{l|c|c}
\hline
& \textbf{Simulation} & \textbf{Reality} \\
\hline
State Space & $\mathbf{u}_k = \mathbf{u}(\mathbf{x}, t_k) \in \mathbb{R}^{n}$ & $\mathbf{u}'_k = \mathbf{u}'(\mathbf{x}, t_k) \in \mathbb{R}^{n}$ \\
Sensor Data & $\mathbf{s}_k = \mathbf{M}\mathbf{u}_k \in \mathbb{R}^{p}$ & $\mathbf{s}'_k \in \mathbb{R}^{p}$ \\
Data Matrix & $\mathbf{X} = [\mathbf{u}_1 \cdots \mathbf{u}_m] \in \mathbb{R}^{n \times m}$ & $\mathbf{X}' = [\mathbf{u}'_1 \cdots \mathbf{u}'_m] \in \mathbb{R}^{n \times m}$ \\
Sensor Matrix & $\mathbf{S} = [\mathbf{s}_1 \cdots \mathbf{s}_m] \in \mathbb{R}^{p \times m}$ & $\mathbf{S}' = [\mathbf{s}'_1 \cdots \mathbf{s}'_m] \in \mathbb{R}^{p \times m}$ \\
Sampling Operator & \multicolumn{2}{c}{$\mathbf{M} \in \mathbb{R}^{p \times n}$} \\
Governing Equations & $\mathbf{u}_t = \mathcal{N}(\mathbf{u}, \mathbf{x}, t)$ & $\mathbf{u}'_t = \mathcal{M}(\mathbf{u}', \mathbf{x}, t)$ \\
\hline
\end{tabular}
\caption{Summary of variables, data, and models used in the Cheap2Rich formulation. The state space is of dimension $n$, there are $m$ snapshots of temporal measurements using $p$ sensors. For reality, only sensor measurements $\mathbf{s}'_k$ are observed; the full state $\mathbf{u}'_k$ is never directly available.}
\label{tab:notation}
\end{table}

\section{High-Frequency Pathway Details}
\label{app:hf_details}

This appendix provides detailed specifications of the high-frequency pathway components described in Section~3.

\subsection{Time-Derivative Embedding}

To capture velocity and acceleration information critical for correcting phase mismatches, the residual history is augmented with finite-difference temporal derivatives:
\begin{align}
\dot{\mathbf{r}}_\ell &= \frac{\mathbf{r}_{\ell+1} - \mathbf{r}_{\ell-1}}{2\Delta t}, \\
\ddot{\mathbf{r}}_\ell &= \frac{\mathbf{r}_{\ell+1} - 2\mathbf{r}_\ell + \mathbf{r}_{\ell-1}}{\Delta t^2},
\label{eq:time_derivatives_app}
\end{align}
forming an augmented input $\tilde{\mathbf{R}} = [\mathbf{r}_\ell, \dot{\mathbf{r}}_\ell, \ddot{\mathbf{r}}_\ell] \in \mathbb{R}^{L \times 3p}$ that embeds the local temporal structure.

\subsection{Temporal Attention Encoder}

The HF encoder employs an attention mechanism over temporal lags to learn which timesteps are most informative for predicting the HF correction:
\begin{align}
\mathbf{H} &= \text{LSTM}_{\text{all}}(\tilde{\mathbf{R}}; \boldsymbol{\theta}_{\text{all}}) \in \mathbb{R}^{L \times d_z}, \\
\mathbf{h}_{\text{final}} &= \text{LayerNorm}\left(\text{LSTM}_{\text{main}}(\tilde{\mathbf{R}}; \boldsymbol{\theta}_{\text{main}})\right) \in \mathbb{R}^{d_z}, \\
\boldsymbol{\alpha} &= \text{softmax}\left(\mathcal{A}(\mathbf{h}_{\text{final}}; \boldsymbol{\theta}_A)\right) \in \mathbb{R}^L, \\
\mathbf{z}_{\text{HF}} &= \sum_{\ell=1}^{L} \alpha_\ell \mathbf{H}_\ell,
\label{eq:hf_encoder_app}
\end{align}
where $\text{LSTM}_{\text{all}}$ returns hidden states at all timesteps, $\text{LSTM}_{\text{main}}$ is a two-layer LSTM returning only the final hidden state, $\mathcal{A}: \mathbb{R}^{d_z} \to \mathbb{R}^L$ is a two-layer attention network with Tanh activation, and $\boldsymbol{\alpha}$ represents learned attention weights over temporal lags.

\subsection{HF Decoder with Spatial Deformation}

The HF decoder generates a base spatial pattern and applies a learned spatially-varying deformation to correct for velocity mismatches:
\begin{equation}
\mathbf{u}_{\text{base}}(t) = \gamma \cdot \mathcal{D}_{\text{HF}}(\mathbf{z}_{\text{HF}}(t); \boldsymbol{\theta}_{\text{HF}}) \in \mathbb{R}^n,
\label{eq:hf_base_app}
\end{equation}
where $\gamma$ is a learnable scale parameter initialized to $0.5$.

To handle spatially and temporally varying wave velocities, we apply a deformation-based correction:
\begin{align}
\boldsymbol{\delta}(t) &= \tanh\left(\mathcal{F}_{\text{shift}}(\mathbf{z}_{\text{HF}}; \boldsymbol{\theta}_{\text{shift}})\right) \cdot \delta_{\max} \in \mathbb{R}^n, \\
\mathbf{a}(t) &= \frac{1}{2}\text{Softplus}\left(\mathcal{F}_{\text{amp}}(\mathbf{z}_{\text{HF}}; \boldsymbol{\theta}_{\text{amp}})\right) + \frac{1}{2} \in \mathbb{R}^n,
\label{eq:deformation_params_app}
\end{align}
where $\boldsymbol{\delta}$ is a position-dependent shift field bounded by $\delta_{\max}$ grid points (typically $\delta_{\max} = 10$), and $\mathbf{a}$ is a positive amplitude modulation field centered around $1$.

The final HF output is obtained via spatial warping with periodic boundary conditions:
\begin{equation}
u_{\text{HF}}(x_i) = a_i \cdot u_{\text{base}}\left((x_i + \delta_i) \mod L_x\right),
\label{eq:warping_app}
\end{equation}
where bilinear interpolation is used for non-integer shifts to ensure smoothness.

\section{Training Pipeline Details}
\label{app:training}

This appendix provides complete details of the four-stage training pipeline for the Cheap2Rich model.

\subsection{Stage 1: SHRED Training on Simulation}

A standard SHRED model is first trained on simulation data with sparse sensor inputs. The model consists of an LSTM that maps sensor histories to a latent space, followed by a decoder MLP:
\begin{equation}
\mathcal{L}_{\text{Stage 1}} = \frac{1}{m} \sum_{k=1}^{m} \left\| \mathcal{D}_{\text{LF}}\left(\mathcal{E}_{\text{LF}}(\mathbf{S}_{t_k-L:t_k})\right) - \mathbf{u}_k \right\|_2^2,
\label{eq:stage1_loss_app}
\end{equation}
where $\mathcal{E}_{\text{LF}}$ denotes the LSTM encoding and $\mathcal{D}_{\text{LF}}$ denotes the decoder. This pretrained SHRED model captures the dominant dynamics of the inexpensive simulation and serves as the LF pathway in the multi-scale architecture.

\subsection{Stage 2: Latent GAN Training}

The generator $\mathcal{G}$ and discriminator $\mathcal{D}$ are trained to align latent distributions:
\begin{equation}
\min_{\boldsymbol{\theta}_G} \max_{\boldsymbol{\theta}_D} \; \mathcal{L}_D + \mathcal{L}_G,
\label{eq:stage2_loss_app}
\end{equation}
where latent codes are extracted from both simulation and real sensor data using the frozen LF encoder.

\subsection{Stage 3: HF-SHRED Training with Spectral Sparsity}

The HF pathway is trained with sensor-only supervision combined with a spectral sparsity regularizer. Let $k_c$ denote a user-specified cutoff frequency. The training objective is:
\begin{equation}
\mathcal{L}_{\text{Stage 3}} = \mathcal{L}_{\text{sensor}} + \lambda \mathcal{R}_{\text{freq}}(\tilde{\mathbf{u}}_{\text{HF}}) + \mu \mathcal{L}_{\text{mag}},
\label{eq:stage3_loss_app}
\end{equation}
where each term is defined as follows.

\paragraph{Sensor Residual Loss.} The HF output must match the observed residual at sensor locations:
\begin{equation}
\mathcal{L}_{\text{sensor}} = \left\| \mathbf{M}\tilde{\mathbf{u}}_{\text{HF}}(t) - \mathbf{r}_t \right\|_2^2.
\label{eq:sensor_loss_app}
\end{equation}

\paragraph{Bandlimited Spectral Sparsity.} Let $\hat{\mathbf{u}}_{\text{HF}}(k)$ denote the discrete Fourier coefficients of $\tilde{\mathbf{u}}_{\text{HF}}$ obtained via rFFT. The regularizer encourages sparsity within a specified frequency band while penalizing energy in higher frequencies:
\begin{equation}
\mathcal{R}_{\text{freq}}(\tilde{\mathbf{u}}_{\text{HF}}) = \underbrace{\frac{\sum_{k=0}^{k_c} |\hat{u}_{\text{HF}}(k)|}{\sqrt{\sum_{k=0}^{k_c} |\hat{u}_{\text{HF}}(k)|^2 + \epsilon}}}_{\text{sparsity within } k \leq k_c} + \beta \underbrace{\frac{\sum_{k > k_c} |\hat{u}_{\text{HF}}(k)|^2}{\sum_{k \geq 0} |\hat{u}_{\text{HF}}(k)|^2 + \epsilon}}_{\text{high-frequency energy ratio}},
\label{eq:sparsity_loss_app}
\end{equation}
where $k_c$ is a user-specified cutoff frequency and $\beta \gg 1$ (typically $\beta = 100$) penalizes energy above this cutoff to encourage sparse, low-frequency corrections.

For the RDE experiments with $p = 25$ sensors over a domain of $n = 100$ grid points, we set the cutoff frequency to $k_c = 12$, which empirically balances reconstruction fidelity with spectral parsimony.

\paragraph{Magnitude Constraint.} To prevent the HF component from dominating:
\begin{equation}
\mathcal{L}_{\text{mag}} = \left[\max\left(0, \|\tilde{\mathbf{u}}_{\text{HF}}\|_1 - \tau\right)\right]^2,
\label{eq:magnitude_loss_app}
\end{equation}
where $\tau$ is estimated from the sensor residual scale.

\paragraph{Warmup Schedule.} The sparsity weight $\lambda$ is ramped from $0$ to its target value over an initial warmup period to allow the model to first learn the basic residual structure before enforcing sparsity.

\subsection{Stage 4: Fine-Tuning}

The HF pathway is fine-tuned with reduced sparsity weight $\lambda' = 0.1\lambda$ to allow greater expressivity while maintaining the learned sparse structure.

\subsection{Inference and Full-State Reconstruction}

At inference time, given a new sensor history $\mathbf{S}'_{t-L:t}$, the full-state estimate is computed as:
\begin{align}
\mathbf{z}_{\text{LF}} &= \mathcal{E}_{\text{LF}}(\mathbf{S}'_{t-L:t}), \\
\tilde{\mathbf{z}}_{\text{LF}} &= \mathbf{z}_{\text{LF}} + \mathcal{G}(\mathbf{z}_{\text{LF}}), \\
\tilde{\mathbf{u}}_{\text{LF}} &= \mathcal{P}_{k_c}\left(\mathcal{D}_{\text{LF}}(\tilde{\mathbf{z}}_{\text{LF}})\right), \\
\mathbf{r}_\ell &= \mathbf{s}'_{t-\ell} - \mathbf{M}\tilde{\mathbf{u}}_{\text{LF}}, \quad \ell = 0, \ldots, L-1, \\
\tilde{\mathbf{u}}_{\text{HF}} &= \text{Deform}\left(\mathcal{D}_{\text{HF}}(\mathcal{E}_{\text{HF}}(\mathbf{R}_{t-L:t}))\right), \\
\tilde{\mathbf{u}}'(t) &= \tilde{\mathbf{u}}_{\text{LF}}(t) + \tilde{\mathbf{u}}_{\text{HF}}(t).
\label{eq:inference_app}
\end{align}

\subsection{Hyperparameter Configuration}
\label{app:hyperparams}

Table~\ref{tab:hyperparams} summarizes the default hyperparameters.
\begin{table}[t]
\centering
\caption{Default hyperparameters for the Sparse-Frequency DA-SHRED architecture.}
\label{tab:hyperparams}
\begin{tabular}{llc}
\toprule
\textbf{Component} & \textbf{Parameter} & \textbf{Value} \\
\midrule
Data & Number of sensors & 25 \\
 & Temporal lags $L$ & 25 \\
 & Train/Valid split & 80\%/20\% \\
\midrule
LF-SHRED Encoder & Hidden dimension $d_z$ & 32 \\
 & LSTM layers & 2 \\
 & Dropout rate & 0.1 \\
 & Normalization & LayerNorm \\
\midrule
LF Decoder & Hidden layers & [128, 128] \\
 & Activation & ReLU \\
\midrule
GAN & Generator hidden & 64 \\
 & Discriminator hidden & 64 \\
 & Activation & LeakyReLU \\
\midrule
HF-SHRED & Hidden dimension & 32 \\
 & LSTM layers & 2 \\
 & Dropout rate & 0.1 \\
 & Lag attention & True \\
 & Attention hidden & 64 \\
 & Velocity correction & Deformation \\
 & Max spatial shift & $\pm 10$ grid points \\
\midrule
HF Decoder & Hidden layers & [128, 128] \\
 & Deformation net & [128, $N$] \\
 & Amplitude net & [64, $N$] \\
 & Scale $\gamma$ init & 0.5 \\
\midrule
Sparsity & $\lambda_{\text{sparse}}$ & 0.1 \\
 & Out-of-band penalty $\beta$ & 100.0 \\
 & Fine-tune $\lambda'_{\text{sparse}}$ & 0.01 \\
\midrule
Optimization & Batch size & 32 \\
 & Optimizer & Adam / AdamW \\
\bottomrule
\end{tabular}
\end{table}

\subsection{Computational Cost and Model Complexity}
\label{app:complexity}

Table~\ref{tab:computation} summarizes the computational cost and model complexity.

\begin{table}[H]
\centering
\caption{Computational cost and model complexity.}
\label{tab:computation}
\begin{tabular}{lc}
\toprule
\textbf{Metric} & \textbf{Value} \\
\midrule
\multicolumn{2}{l}{\textit{Model Complexity}} \\
\quad LF-SHRED parameters & 4.9 K \\
\quad HF-SHRED parameters & 163.9 K \\
\quad Total parameters & 168.8 K \\
\midrule
\multicolumn{2}{l}{\textit{Training Time}} \\
\quad Stage 1 (LF-SHRED) & 7.92 sec \\
\quad Stage 2 (GAN) & 1.09 sec \\
\quad Stage 3 (HF-SHRED + Fine-tuning) & 38.87 sec \\
\quad Total & 47.88 sec \\
\midrule
\multicolumn{2}{l}{\textit{Hardware}} \\
\quad CPU & Apple M4 Pro \\
\quad Memory & 24 GB \\
\bottomrule
\end{tabular}
\end{table}

\newpage
\section{Comparison to other methods}
\label{app:baselines}

This appendix benchmarks our approach against standard supervised baselines. We assume access to paired low-fidelity and high-fidelity datasets, and consider the reconstruction task
\[
\mathbb{R}^p \ni s_t \;\mapsto\; u'_t \in \mathbb{R}^n,
\]
where $s_t$ denotes the $p$ sensor measurements at time $t$ and $u'_t$ the corresponding high-fidelity state. We compare (i) SHRED models that use a history of sensor measurements (lagged inputs) and (ii) feed-forward MLPs that map sensors to state instantaneously. For each architecture, we evaluate two training strategies: training solely on high-fidelity data, and pretraining on low-fidelity data followed by fine-tuning on high-fidelity data.

All models are evaluated with an RMSE calculated on the final 50 snapshots of the high-fidelity dataset, held out from training, as well as SSIM evaluated at the entire high-fidelity dataset. When pretraining is used, we train on 250 preprocessed snapshots from Koch's model before fine-tuning on the high-fidelity data.

\subsection{SHRED trained on high-fidelity data only}
\begin{figure}[H]
  \centering
  \includegraphics[width=\linewidth]{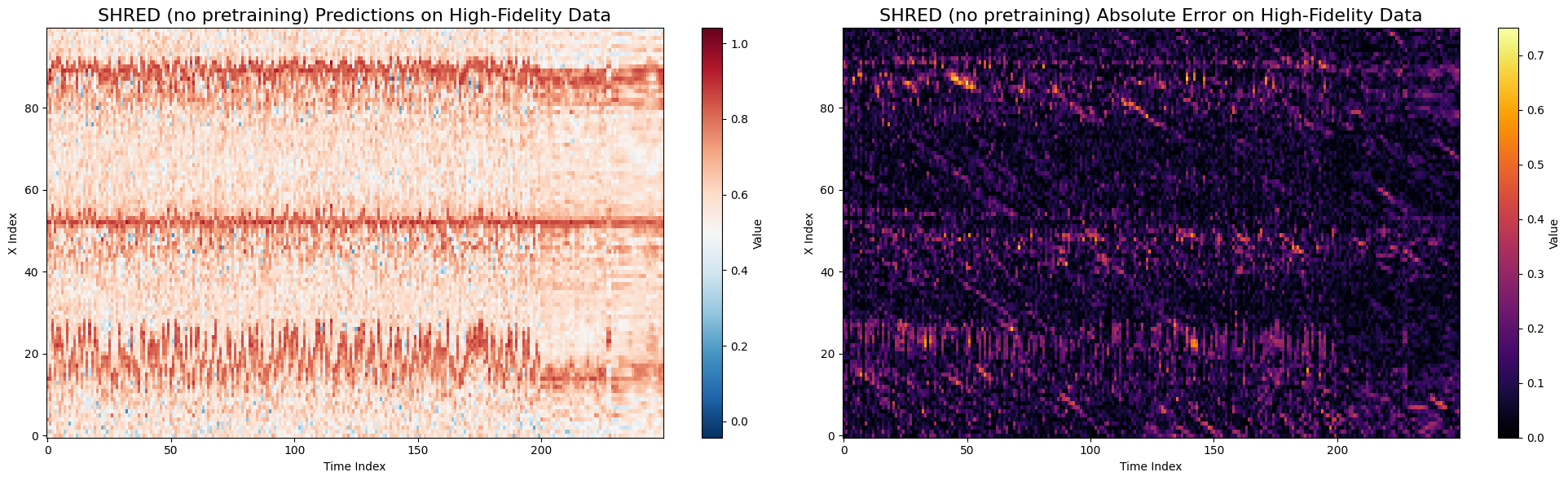}
  \caption{SHRED trained on high-fidelity data only.}
  \label{fig:SHRED_no_p}
\end{figure}

We use $p=25$ sensors, $L=5$ lags, and a latent dimension $d_z=32$. The model is trained for 500 epochs with Adam (initial learning rate $10^{-3}$), achieving $\mathrm{RMSE}=0.102237$ on the test set and SSIM of $0.171135
$ on train + test. To enable inference for $t \le L$, we pad the sensor histories in both the training and test sets with zeros at the initial time steps. Reconstructions are shown in Figure~\ref{fig:SHRED_no_p}.

\subsection{SHRED pretrained on Koch's model, then fine-tuned on high-fidelity data}
\begin{figure}[H]
  \centering
  \includegraphics[width=\linewidth]{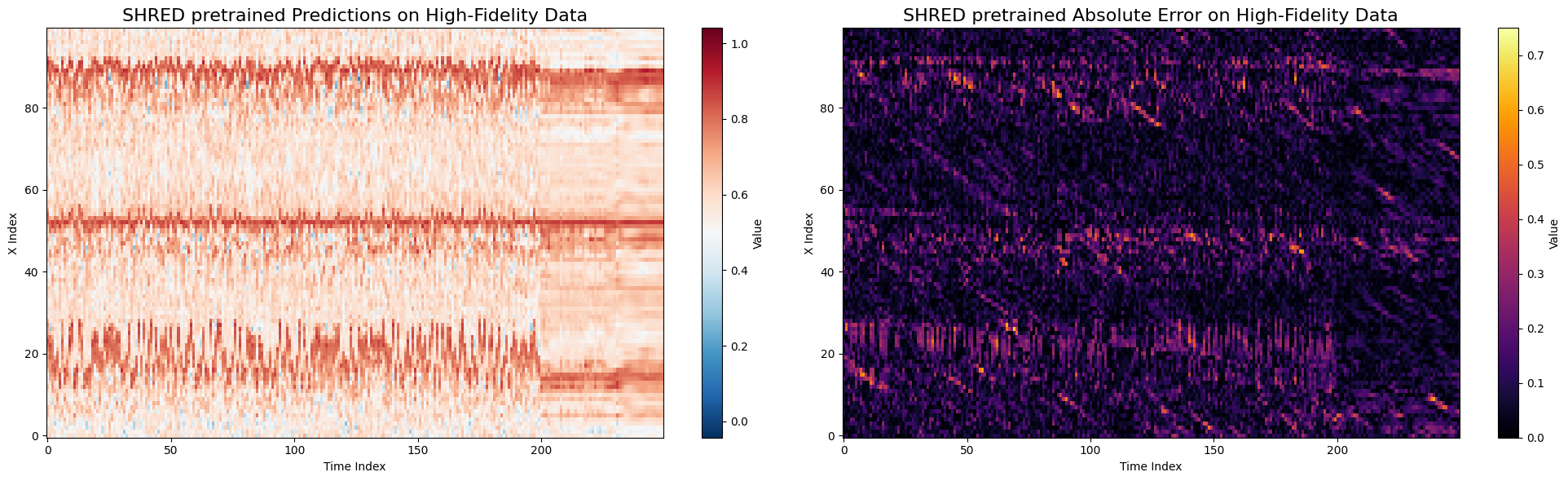}
  \caption{SHRED pretrained on Koch's data and fine-tuned on high-fidelity data.}
  \label{fig:SHRED_p}
\end{figure}

We again use $p=25$, $L=5$, and $d_z=32$. The model is pretrained for 500 epochs on Koch's data and then fine-tuned for 300 epochs on the high-fidelity data using Adam (initial learning rate $10^{-3}$). This yields $\mathrm{RMSE}=0.107945$ on the test set, and SSIM of $0.176757$ on train + test. As above, sensor histories are zero-padded to support inference for $t \le L$. Reconstructions are shown in Figure~\ref{fig:SHRED_p}.

\subsection{MLP trained on high-fidelity data only}
\begin{figure}[H]
  \centering
  \includegraphics[width=\linewidth]{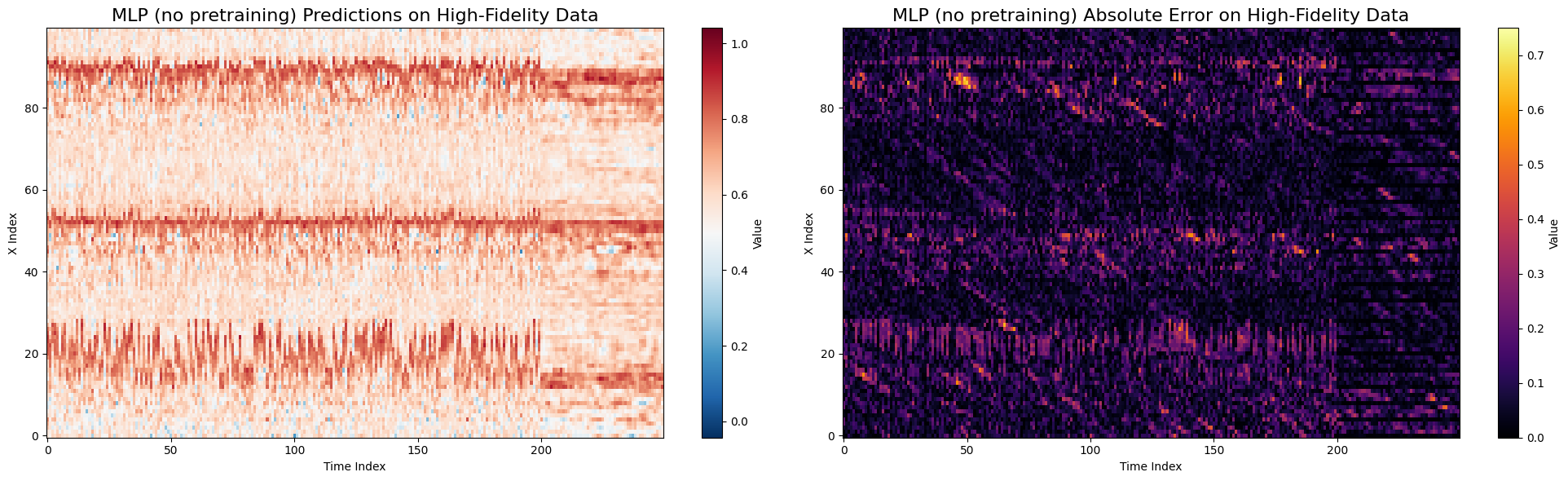}
  \caption{MLP trained on high-fidelity data only.}
  \label{fig:MLP_no_p}
\end{figure}

We use $p=25$ sensors and a feed-forward MLP with layer widths $[25,128,128,128,100]$ and ReLU activations. The model achieves $\mathrm{RMSE}=0.093735$ on the test set and SSIM of $0.213414$ on train + test. Reconstructions are shown in Figure~\ref{fig:MLP_no_p}.

\subsection{MLP pretrained on Koch's model, then fine-tuned on high-fidelity data}
\begin{figure}[H]
  \centering
  \includegraphics[width=\linewidth]{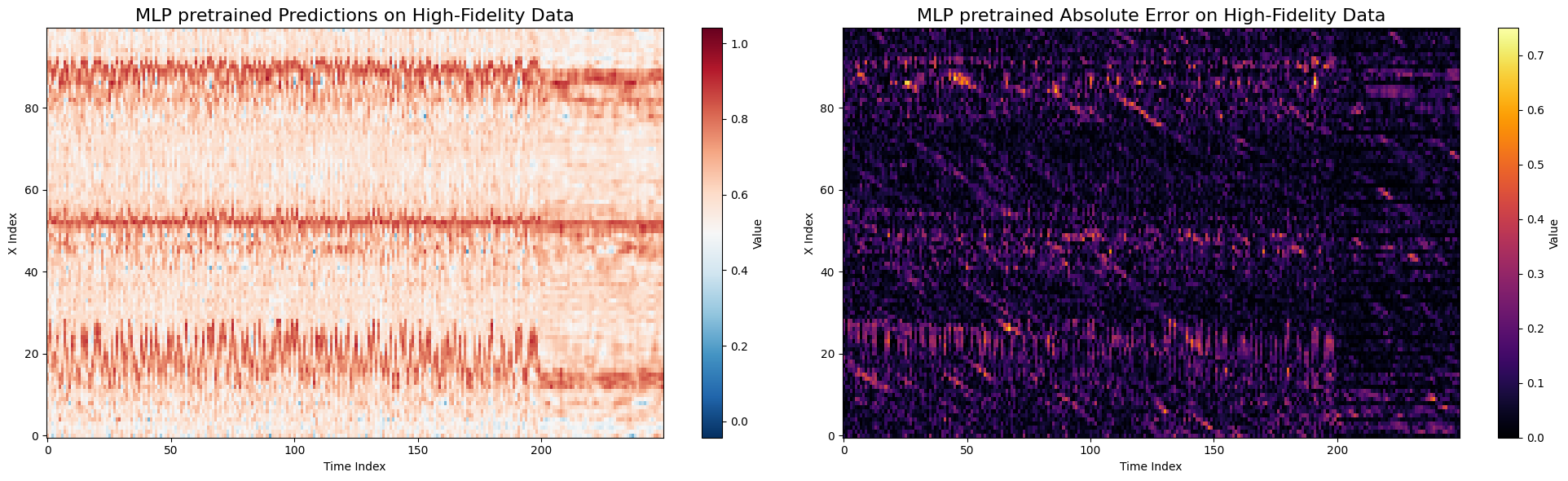}
  \caption{MLP pretrained on Koch's data and fine-tuned on high-fidelity data.}
  \label{fig:MLP_p}
\end{figure}

We use the same MLP architecture as above. The model is pretrained for 500 epochs on Koch's data and then fine-tuned for 300 epochs on the high-fidelity data with Adam (initial learning rate $10^{-3}$), achieving $\mathrm{RMSE}=0.107945$ on the test set, and SSIM of $0.213414$ on train + test. Reconstructions are shown in Figure~\ref{fig:MLP_p}.

\subsection{Cheap2Rich (ours)}
\begin{figure}[H]
  \centering
  \includegraphics[width=\linewidth]{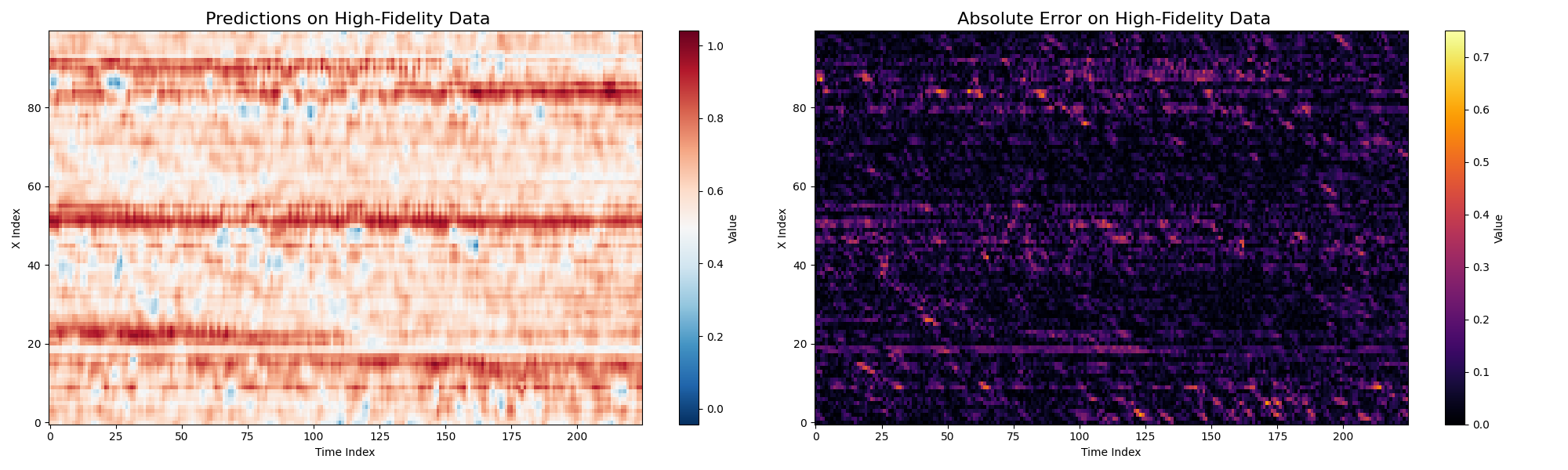}
  \caption{Cheap2Rich reconstructions on the high-fidelity test set.}
  \label{fig:ours}
\end{figure}

Figure~\ref{fig:ours} reports reconstructions produced by Cheap2Rich. The test-set error is $\mathrm{RMSE}=0.1031$, and the SSIM on the entire high-fidelity dataset is $0.3638$. While the aggregate RMSE is comparable to the baselines, the reconstructions better preserve the salient spatiotemporal structure, which is also explained by the much higher SSIM. In particular, the three dominant wavefronts are sharper and more consistently separated, and fine-scale features are recovered more faithfully, including the colder regions near $(x,t)=(80,25)$ and $(45,75)$.

\section{SINDy Framework and Simulation Integration Details}
\label{app:sindy}

This appendix provides additional details on the SINDy-based physics discovery described in Section~5.

\subsection{SINDy Framework for Discrepancy Modeling}

Given the LF reconstruction $\tilde{\mathbf{u}}_{\text{LF}}$, the HF correction $\tilde{\mathbf{u}}_{\text{HF}}$, and the known Koch's simulation $\mathbf{u}_{\text{sim}}$, we can pose three complementary discovery problems:

\paragraph{LF Correction Dynamics.} The GAN-based latent alignment induces a correction to the simulation output. We seek to identify the functional form of this correction:
\begin{equation}
\frac{\partial}{\partial t}\left(\tilde{\mathbf{u}}_{\text{LF}} - \mathbf{u}_{\text{sim}}\right) = f\left(\Delta u, \Delta u_x, u_{\text{LF}}, u_{\text{LF},x}, \ldots\right),
\label{eq:lf_correction_app}
\end{equation}
where $\Delta u = \tilde{\mathbf{u}}_{\text{LF}} - \mathbf{u}_{\text{sim}}$ denotes the LF correction field.

\paragraph{HF Component Dynamics.} The high-frequency pathway captures fine-scale physics absent from the Koch's model. We identify its governing dynamics:
\begin{equation}
\frac{\partial \tilde{\mathbf{u}}_{\text{HF}}}{\partial t} = g\left(u_{\text{HF}}, u_{\text{HF},x}, u_{\text{LF}}, u_{\text{LF}} \cdot u_{\text{HF}}, \ldots\right).
\label{eq:hf_dynamics_app}
\end{equation}

\paragraph{Direct Missing Physics.} Most directly, we can identify the total discrepancy between real and simulated dynamics:
\begin{equation}
\frac{\partial}{\partial t}\left(\mathbf{u}_{\text{real}} - \mathbf{u}_{\text{sim}}\right) = h\left(u, u_x, u_{xx}, u \cdot u_x, \ldots\right).
\label{eq:direct_missing_app}
\end{equation}
where $ u $ denotes $\mathbf{u}_{\text{sim}}$.

\subsection{Library Construction and Sparse Regression}

For each discovery problem, we construct a library of candidate nonlinear terms $\boldsymbol{\Theta}$ and solve the sparse regression problem:
\begin{equation}
\frac{\partial \mathbf{u}}{\partial t} = \boldsymbol{\Theta}(\mathbf{u}, \mathbf{u}_x, \mathbf{u}_{xx}, \ldots) \boldsymbol{\xi},
\label{eq:sindy_regression_app}
\end{equation}
where $\boldsymbol{\xi}$ is a sparse coefficient vector recovered via sequential thresholded least squares (STLSQ) \cite{brunton2016discovering}.

The candidate library is tailored to the physical context of RDE dynamics. For the direct missing physics discovery, we use:
\begin{equation}
\boldsymbol{\Theta} = \left[1, \; u, \; u^2, \; u^3, \; u_x, \; u_{xx}, \; u \cdot u_x, \; \Delta u, \; \Delta u_x, \; \ldots \right],
\label{eq:library_app}
\end{equation}
where spatial derivatives are computed spectrally via FFT to ensure accuracy \cite{brunton2022data}. The library is normalized column-wise before regression to ensure fair coefficient comparison across terms of different magnitudes.

The STLSQ algorithm iteratively performs least-squares regression and thresholds small coefficients to zero, promoting sparsity while maintaining accuracy. We employ adaptive thresholding, starting from $\alpha = 0.001$ and incrementing until the discovered equation contains 2--8 active terms, balancing parsimony with expressiveness.

\subsection{Simulation Integration}

The discovered equations provide two pathways for improving the Koch's model without requiring the neural network at inference time.

\paragraph{Direct Correction.} The most straightforward approach uses the discovered missing physics equation directly. The modified governing equation becomes:
\begin{equation}
\frac{\partial u}{\partial t} = \left[\text{Koch's terms}\right] + \gamma \cdot h(u, u_x),
\label{eq:modified_koch_direct_app}
\end{equation}
where $h(u, u_x) = 0.72 - 3.79u + 6.24u^2 - 3.33u^3 + 0.40 u_x - 0.40 u \cdot u_x$ is the discovered correction functional and $\gamma$ is a tunable scale factor that accounts for normalization differences between training data and simulation variables.

\paragraph{Hierarchical Correction.} Alternatively, the LF and HF corrections can be applied separately, enabling independent tuning of large-scale adjustments versus fine-scale dynamics:
\begin{equation}
\frac{\partial u}{\partial t} = \left[\text{Koch's terms}\right] + \gamma_{\text{LF}} \cdot f(\Delta u, u) + \gamma_{\text{HF}} \cdot g(u_{\text{HF}}, u_{\text{LF}}),
\label{eq:modified_koch_hierarchical_app}
\end{equation}
where $f$ and $g$ are the discovered LF correction and HF dynamics, respectively. This approach provides finer control but requires tracking both components during simulation.

\subsection{Connection to DA-SHRED Framework}

The SINDy-based discovery presented here extends the methodology of \cite{bao2025data} to the multi-scale setting. While the original DA-SHRED framework demonstrated discrepancy modeling for systems with a single learned correction, the multi-scale architecture provides a natural decomposition that yields richer physical insight:

\begin{itemize}
\item The \textbf{LF correction} captures what the latent-space GAN alignment contributes---primarily amplitude and baseline adjustments that align the simulation manifold with reality.

\item The \textbf{HF dynamics} reveal the structure of physics entirely absent from the simplified model, including the coupling between fine-scale fluctuations and the dominant wave structure.

\item The \textbf{direct missing physics} provides a single equation summarizing all corrections needed, suitable for integration into simulations where the model discrepancy is perturbative---i.e., when the simulation is not far from reality.
\end{itemize}

This combined discovery approach---enabled by the Cheap2Rich architecture---provides both interpretability and actionable model corrections, advancing beyond pure reconstruction toward genuine physics discovery.




\end{document}